\documentclass[twoside]{article}
\usepackage[accepted]{aistats2025}
\usepackage[round]{natbib}

\usepackage{acro}
\usepackage{xfrac}

\newcommand{\reals}{\mathbb{R}}

\newcommand{\norm}[1]{\|#1\|}

\newcommand{\numberthis}{\addtocounter{equation}{1}\tag{\theequation}}

\newcommand{\gramian}{G}
\newcommand{\itermat}{M}
\newcommand{\ccitermat}{\tilde{M}}
\newcommand{\itervec}{g}

\newcommand{\gramdiag}{D_\gramian}
\DeclareAcronym{GP}{short = GP, long = Gaussian process}
\DeclareAcronym{PSIM}{short = PSIM, long = probabilistic stationary iterative method}
\DeclareAcronym{PLS}{short = PLS, long = probabilistic linear solver}
\DeclareAcronym{PNM}{short = PNM, long = probabilistic numerical method}
\DeclareAcronym{CG}{short = CG, long=the conjugate gradient method}

\DeclareAcronym{CAGP}{
    short = CAGP ,
    long  = computation-aware GP
}
\DeclareAcronym{RMSE}{short = RMSE, long=root mean square error}
\DeclareAcronym{NLL}{short = NLL, long=negative log likelihood}

\usepackage{pdfpages}
\usepackage{enumitem}
\usepackage{amsmath, amsthm, amssymb}
\usepackage{subcaption}
\usepackage{hyperref}
\usepackage{algorithm}
\usepackage{algorithmicx,algpseudocodex}

\usepackage{xr-hyper}

\usepackage[capitalise]{cleveref}

\newtheorem{proposition}{Proposition}

\newtheorem{theorem}[proposition]{Theorem}
\newtheorem{corollary}[proposition]{Corollary}
\newtheorem{definition}[proposition]{Definition}

\begin{document}

\twocolumn[

\aistatstitle{Calibrated Computation-Aware Gaussian Processes}

\aistatsauthor{ Disha Hegde \And Mohamed Adil \And Jon Cockayne }

\aistatsaddress{ University of Southampton \And  \And University of Southampton } ]

\begin{abstract}
Gaussian processes are notorious for scaling cubically with the size of the training set, preventing application to very large regression problems.
Computation-aware Gaussian processes (CAGPs) tackle this scaling issue by exploiting probabilistic linear solvers to reduce complexity, widening the posterior with additional \emph{computational} uncertainty due to reduced computation.
However, the most commonly used CAGP framework results in (sometimes dramatically) conservative uncertainty quantification, making the posterior unrealistic in practice.
In this work, we prove that if the utilised probabilistic linear solver is \emph{calibrated}, in a rigorous statistical sense, then so too is the induced CAGP.
We thus propose a new CAGP framework, CAGP-GS, based on using Gauss-Seidel iterations for the underlying probabilistic linear solver.
CAGP-GS performs favourably compared to existing approaches when the test set is low-dimensional and few iterations are performed.
We test the calibratedness on a synthetic problem, and compare the performance to existing approaches on a large-scale global temperature regression problem.
\end{abstract}

\section{INTRODUCTION}

Gaussian processes are a powerful and flexible tool for Bayesian nonparametric regression, allowing a user to fit a wide array of possibly nonlinear phenomena using simple computational routines.
The most major challenge in scaling Gaussian process regression to high-dimensional datasets is its cubic scaling with the number of training points, arising from the need to invert a Gramian  matrix representing the prior covariance between data points \citep{rasmussen_gaussian_2005}.
To address this, a wide array of computational approximations have been proposed, including iterative solvers (e.g.\ \cite{Wenger2022precond}), approximations to the kernel matrix (e.g.\ \cite{ferrari-trecate_finite-dimensional_1998}), and inducing point methods (e.g.\ \cite{Titsias2009Variational}).

A particularly appealing approach recently proposed in \cite{Wenger2022Computational} is \acp{CAGP}.
In this framework one uses a \ac{PLS} to solve a linear system involving the Gramian matrix and then marginalises the uncertainty from the \ac{PLS}.
The result is an elegant cancellation that eliminates the need to invert the Gramian (see \cref{sec:marginalisation} for a more detailed explanation).
The new posterior obtained is called ``computation aware'' because it is widened to represent additional \emph{computational uncertainty} due to reduced computation in the \ac{PLS}, compared to an exact (but computationally prohibitive) linear solve.

The most commonly used \ac{PLS} for \acp{CAGP} is BayesCG (see \cite{Cockayne2019BayesCG})---we refer to this setting as \ac{CAGP}-CG.
This approach is favoured because of rapid mean convergence.
One of the major challenges with \ac{CAGP}-CG is that resulting \ac{CAGP} is typically conservative---the posterior mean is much closer to the truth than the width of the posterior covariance suggests it should be.
This is inherited from the widely observed poor \emph{calibration} properties of BayesCG, described in several works including \cite{Cockayne2019BayesCG,Bartels2019Unifying,wenger2020problinsolve,reid_statistical_2023}.
We formally introduce calibration in \cref{sec:calib}.
While some empirical Bayesian methods for mitigating this issue have been proposed (e.g., \cite{wenger2020problinsolve,reid_bayescg_2022,reid_statistical_2023}), they are difficult to apply within \acp{CAGP} as the choice of prior is strictly constrained by the method.

\subsection{Contributions}

The contributions of this paper are as follows:
\begin{itemize}[nosep]
    \item We rigorously prove that, if a calibrated \ac{PLS} is used, the resulting \ac{CAGP} is calibrated (\cref{thm:itergp_calib}).
    \item We introduce a new class of \acp{CAGP} based on \acp{PSIM} (as seen in \cite{Cockayne2021PIMs}).
    \item We explore efficient implementation of a new \ac{CAGP} approach based on Gauss-Seidel iterations (\ac{CAGP}-GS, \cref{sec:GaussSeidal}).
    \ac{CAGP}-GS scales favourably compared to \ac{CAGP}-CG in regimes where only a small number of test points are required.
    \item We explore the empirical properties of \ac{CAGP}-GS on a synthetic test problem, some benchmark datasets and a large-scale geospatial regression problem (\cref{sec:synthetic,sec:benchmark,sec:era5}).
    In particular, we note that for small iteration numbers, \ac{CAGP}-GS outperforms all known alternative \ac{CAGP} frameworks in terms of mean convergence and uncertainty quantification.
\end{itemize}

\subsection{Structure of the Paper}

The rest of the paper proceeds as follows.
In \cref{sec:itergp} we discuss the required background on \acp{CAGP}, \acp{PLS} and calibratedness. \cref{sec:calib_cagp} discusses calibratedness in the context of \acp{CAGP}, while \cref{sec:inverse_pims} presents a particular class of \acp{PLS} that are both calibrated and can be used in the \ac{CAGP} framework.
\cref{sec:simulations} presents simulations demonstrating the new methodology, and we conclude in \cref{sec:conclusion}.
Proofs and additional results for experiments from \cref{sec:simulations} are included in the supplementary material.

\section{BACKGROUND} \label{sec:itergp}

\subsection{GP Regression} \label{sec:GPs}

Suppose $f \sim \mu_0 = \mathcal{GP}(m_0, k)$, where $k$ is a positive definite kernel function.
Consider \ac{GP} inference under the observation model
\begin{equation}
    y = f(X) + \zeta
\end{equation}
where $X \subset D$ is a set of $d$ distinct training points in the domain of $f$, while $\zeta \sim \mathcal{N}(0, \sigma^2 I)$.
It may be helpful to think of $D$ as a subset of $\reals^n$ for some $n$, but this is not required.
The predictive distribution conditional on this information at a set of $d_\textsc{test}$ test points $X'$ is given by
\begin{subequations}
\begin{align}
    f(X') \mid y &\sim \mu = \mathcal{GP}(\bar{m}, \bar{k}) \\
    \bar{m}(X') &= m_0(X') + k(X', X) v^\star \label{eq:reparam_mean} \\
    \bar{k}(X', X') &= k(X', X') - k(X', X) \gramian^{-1} k(X, X') \label{eq:reparam_cov}
\end{align} \label{eq:gp_posterior}
\end{subequations}
where $\gramian = K(X, X) + \sigma^2 I$, while $v^\star$ is the solution to the linear system
\begin{equation} \label{eq:the_system}
    \gramian v^\star = b.
\end{equation}
with $b = (y - m_0(X))$.

As has been widely noted (e.g., in \cite{rasmussen_gaussian_2005}), one of the major challenges of the \ac{GP} regression is that the complexity of the computations above is $\mathcal{O}(d^3)$ owing to the inversion of the Gramian matrix $G$.
Recent work (\cite{Wenger2022Computational}) proposes a novel framework to mitigate this cost called \acp{CAGP}, introduced next.

\subsection{Computation-Aware GPs} \label{sec:CAGPs}

\acp{CAGP} center on use of \acp{PLS} to solve the system \cref{eq:the_system}.
We will first outline the literature on \acp{PLS} before discussing \acp{CAGP} themselves in \cref{sec:marginalisation}.

\subsubsection{(Bayesian) PLS} \label{sec:BPLS}

At the most generic level, \acp{PLS} are \acp{PNM} (\cite{hennig_probabilistic_2022,cockayne_bayesian_2019}) for solving linear systems, that is, they are learning procedures that return a probability distribution intended to quantify error due to having expended reduced computational effort\footnote{e.g.\ compared to applying a direct method such as Cholesky factorisation followed by two triangular solves, which calculates $v^\star$ precisely in exact arithmetic, but have cubic complexity.} to calculate $v^\star$.
It is common for such learning procedures to depend on some ``prior'' belief about $v^\star$, expressed through the distribution $\eta_0$. We therefore use the notation $\eta : \mathcal{P}(\reals^d) \times \reals^d \to \mathcal{P}(\reals^d)$ or $(\eta_0, b) \mapsto \eta(\eta_0, b)$, where $\mathcal{P}(\reals^d)$ is the set of all probability measures on $\reals^d$ for a \ac{PLS}.
We will limit attention to Gaussian learning procedures, i.e.\ those that accept Gaussian input $\eta_0 = \mathcal{N}(m_0, C_0)$ and return a Gaussian output $\eta(\eta_0, b) = \mathcal{N}(\bar{v}, \bar{C})$.
As we see in \cref{sec:marginalisation}, the particular prior $\eta_0 = \mathcal{N}(0, G^{-1})$ (referred to as the ``inverse prior'' in this paper) is intrinsic to \acp{CAGP}, and we will limit attention to this throughout the following sections.
As will be described in \cref{sec:marginalisation}, use of this prior is a fundamental requirement for \acp{CAGP} as currently understood.

Many \acp{PLS} have a Bayesian interpretation, i.e.\ they are based on conditioning $\eta_0$ on observations of the form $z_m = S_m^\top b$, where $S_m \in \reals^{d \times m}$ is a matrix of \emph{search directions} with linearly independent columns.
Under such observations and the inverse prior, the posterior is
\begin{subequations}
\begin{align}
    v \mid z_m = \eta &\sim \mathcal{N}(\bar{v}_m, \bar{C}_m) \\
    \bar{v}_m &=   S_m (S_m^\top \gramian  S_m)^{-1} S_m^\top b \label{eq:pls_mean} \\
    \bar{C}_m &= \gramian^{-1} - D_m^{\textsc{Bayes}} \label{eq:pls_cov} \\
    \bar{D}_m^{\textsc{Bayes}} &= S_m (S_m^\top \gramian S_m)^{-1} S_m^\top \label{eq:pls_dd}.
\end{align} \label{eq:pls_posterior}
\end{subequations}
This posterior is not directly computable, as computing $\bar{C}_m$ requires computation of $\gramian^{-1}$, which we assumed in \cref{sec:GPs} we did \emph{not} want to compute.
Nevertheless, as we will see in \cref{sec:marginalisation}, this choice leads to some cancellations for \acp{CAGP}.

\paragraph{Choice of Search Directions}
Generic choices of $S_m$ that have been examined in the literature include standard Euclidean basis vectors and random unit vectors \citep{Cockayne2019BayesCG,Wenger2022Computational,Pfortner2024CAKF}.
However, these typically suffer from slow convergence of $v_m \to v^\star$ compared to state-of-the-art iterative methods, making them unattractive.

A particularly important choice of $S_m$ are those based on \ac{CG}\footnote{Often these are instead obtained from the Lanczos algorithm, which provides directions that span the same space but have different orthogonality properties.}.
In this case the \ac{PLS} is often referred to as BayesCG \citep{Cockayne2019BayesCG}.
These directions are favoured because (i) they can be proven to converge at a geometric rate in $m$ in the worst case (often faster in practice), and (ii) they are $\gramian$-conjugate, meaning that $S_m^\top \gramian S_m$ is diagonal.
As a result the posterior reported in \cref{eq:pls_posterior} simplifies further.
On the other hand, these directions result in poor calibration of the posterior, which will be discussed further in \cref{sec:calib}.

\paragraph{Computational Complexity}
Ignoring the cost of computing $\gramian^{-1}$,
the complexity of computing \cref{eq:pls_mean,eq:pls_dd} is $\mathcal{O}(m (m^2 + d^2))$ owing to the inversion of the $m \times m$ matrix $S_m^\top \gramian S_m$ and the requirement to compute (dense) products of the form $\gramian S_m$ which have complexity $\mathcal{\mathcal{O}}(m d^2)$.
Thus, if $m \ll d$ the complexity compared to a direct method is significantly reduced.
For BayesCG the diagonal matrix inversion is reduced to $\mathcal{O}(m)$, so that the complexity is only $\mathcal{O}(m d^2)$.

\subsubsection{The Marginalisation Trick} \label{sec:marginalisation}

The central ``trick'' behind \acp{CAGP} is that, given any belief from a \acp{PLS} of the form
\begin{equation} \label{eq:cagp_req_posterior}
v \sim \eta = \mathcal{N}(\bar{v}, \gramian^{-1} - D)
\end{equation}
we can construct a new belief over $f(X')$ through the marginalisation
\begin{equation}
    \tilde{p}(f) = \int_{\reals^d} p(f(X') \mid v) \eta(\mathrm{d}v)
\end{equation}
the law of which we denote $\tilde{\mu}$.
Since all the involved distributions are Gaussian, \cite{Wenger2022Computational} derived the modified posterior as
\begin{align}
    \tilde{\mu} &= \mathcal{GP}(\tilde{m}, \tilde{k}) \\
    \tilde{m}(X') &= m_0(X') + k(X', X) \bar{v} \label{eq:cagp_mean}\\
    \tilde{k}(X', X') &= k(X', X') - k(X', X) D k(X, X') \label{eq:cagp_cov}.
\end{align}
Notably, the presence of $\gramian^{-1}$ in $\eta$ is \emph{crucial} as it leads to cancellation of the downdate involving $G^{-1}$ from \cref{eq:reparam_cov}.
As a result we need only calculate the $D$ term in the underlying PLS, negating any need to invert $G$.
If $D$ can be computed at significantly lower complexity than $G^{-1}$, as in \cref{sec:BPLS}, then the overall cost of GP inference is reduced.
Moreover, \citet[Section 2]{Wenger2022Computational} demonstrated that $\tilde{k}(X', X')$ is ``wider'' than $\bar{k}(X', X')$ and can be interpreted as providing additional uncertainty quantification for the reduced computation.

\begin{algorithm}
    \caption{Computation Aware GP} \label{alg:cagp}
    \begin{algorithmic}[1]
    \Function{\textsc{cagp}}{$X$, $y$, $X'$, $\sigma^2$}
    \State $b = y - m_0(X)$
    \State $V = k(X', X)$
    \State $G(v) = v \mapsto k(X, X) v + \sigma^2 v$
    \State $\tilde{v}, \tilde{D} = \textsc{cagp\_pls}(G, b, V)$
    \State $\tilde{m} = m_0(X') + \tilde{v}$
    \State $\tilde{k} = k(X', X') - \tilde{D}$
    \State \Return $\tilde{m}, \tilde{k}$
    \EndFunction
    \end{algorithmic}
\end{algorithm}

\cref{alg:cagp} gives an implementation of this as pseudocode.
Note that it is assumed $\textsc{cagp\_pls}$ requires access to $G$ only through its action on vectors $v$, rather than explicitly, allowing for a matrix-free implementation.
\cref{alg:cagp} is a slightly modified version of that presented in \cite{Wenger2022Computational}; the quantities $\tilde{v}$ and $\tilde{D}$ returned by the routine \textsc{cagp\_pls} are given by $\tilde{v} = V \bar{v} = k(X', X) \bar{v}$, and $\tilde{D} = V D V^\top = k(X', X) D k(X, X')$, i.e. they are the image of $\eta$ under the map $v \mapsto V v$.
This is more commensurate with the novel algorithms we will introduce in \cref{sec:calib_cagp,sec:inverse_pims}.

\subsection{Calibrated Learning Procedures} \label{sec:calib}

While BayesCG has several numerically appealing properties, as already mentioned it is \emph{poorly calibrated}.
In particular the uncertainty quantification provided is conservative, meaning that the posterior covariance $\bar{C}_m$ is typically much wider than the error $\bar{v}_m - v^\star$.

This is due to an incorrect application of Bayesian inference in constructing the posterior.
It can be shown \citep[Section 11.3.3]{GVL} that (when the initial guess for the solution is zero) the \ac{CG} directions form a basis of the Krylov subspace
\begin{align*}
    K_m(\gramian, b) &=\textup{span}(b, Gb, G^2b, \dots, G^{m-1}b) \\
    &=\textup{span}(G v^\star, G^2 v^\star, \dots, G^m v^\star).
\end{align*}
If we ignore the orthogonalisation of the search directions (which would not affect the posterior in a Bayesian inference problem) this results in information of the form $z_i = s_i^\top G v^\star = (v^\star)^\top G^{i+1} v^\star$, $i=1,\dots,m$, each of which is quadratic in $v^\star$ rather than linear.
Linearity of the information $z_i$ is intrinsic to the Gaussian conditioning argument that underpins \acp{PLS}, and ignoring this yields the poor calibration of BayesCG-based \acp{PLS}.

In this paper we will discuss how another class of \acp{PLS} can be used in the \ac{CAGP} framework.
These \acp{PLS} do not have a Bayesian interpretation, but can nevertheless be said to be \emph{calibrated} in a formal sense.
We will be interested in the question of whether, when the \ac{PLS} is calibrated, the derived \ac{CAGP} is also calibrated, so we present this rather generically.
\cite{Cockayne2022Calib} introduces the notion of \emph{strong calibration}, a more intuitive description of which is given below; we refer the reader to the aforementioned paper for a formal introduction.

Briefly, a learning procedure $\mu : \mathcal{P}(\mathcal{U}) \times \mathcal{Y} \to \mathcal{P}(\mathcal{U})$ is said to be strongly calibrated with respect to a distribution $\mu_0$ and a data-generating model $\textsc{dgm} : \mathcal{U} \to \mathcal{Y}$ if, under the following procedure:
\begin{enumerate}
    \item $u^\star \sim \mu_0$
    \item $y = \textsc{dgm}(u^\star)$
\end{enumerate}
it holds that, on average over $\mu_0$, $u^\star$ is a ``plausible sample'' from the posterior $\mu(\mu_0, y)$.
The last statement can be made formal in several ways, but a particularly simple definition for the case of Gaussian learning procedures is given by \cite{Cockayne2021PIMs}.

\begin{definition}[{\citet[Definition 6 and 9]{Cockayne2021PIMs}}] \label{def:calib}
    Consider a fixed Gaussian prior $\mu_0 = \mathcal{N}(u_0, \Sigma_0)$, a data-generating model $\textup{\textsc{dgm}}$ and a learning procedure $\mu(\mu_0, y) = \mathcal{N}(\bar{u}, \bar{\Sigma})$.
    Suppose that $\bar{\Sigma}$ is independent of $y$ and potentially singular, with $N$ and $R$  matrices whose columns form (mutually) orthonormal bases of its null and row spaces respectively.
    Then $\mu$ is said to be \emph{strongly calibrated} to $(\mu_0, \textup{\textsc{dgm}})$ if:
    \begin{enumerate}
        \item $(R^\top \bar{\Sigma} R)^{-\frac{1}{2}} R^\top (\bar{u} - u^\star) \sim \mathcal{N}(0, I)$.
        \item $N^\top(\bar{u} - u^\star) = 0$
    \end{enumerate}
    when $u^\star \sim \mu_0$.
\end{definition}

Note that in \cref{def:calib} the randomisation of $u^\star$ induces randomness in $\bar{u}$ through dependence on $y$, making the statement rather nontrivial.
The assumption that $\bar{\Sigma}$ is independent of $y$ is important to ensure that the range and null spaces are consistent across draws from $\mu_0$; this can be relaxed, but resulting definitions are far less analytically tractable\footnote{Note however that this precludes selecting hyperparameters using empirical Bayesian procedures.}.
Lastly in the case that $\bar{\Sigma}$ is full rank, the second of the two conditions is redundant.

\cref{def:calib} provides a theoretical framework for validating strong calibration, but we are also interested in validating this numerically.
To accomplish this, we will apply the simulation-based calibration tests of \cite{Talts2018}; these are described in more detail in \cref{sec:sbc}.

\subsection{Probabilistic Stationary Iterative Methods} \label{sec:PIMs}

Another subclass of \acp{PLS} are \acp{PSIM}, introduced in \cite{Cockayne2021PIMs}\footnote{These were originally termed ``probabilistic iterative methods'', but we adopt different nomenclature to avoid confusion (since the methods described in \cref{sec:BPLS} are also both probabilistic and iterative).}.
\acp{PSIM} are based on an underlying \emph{stationary iterative method} for solving the linear system (see e.g.\ \cite{Young1971}), that is, methods that evolve an iterate $v_m$ according to the map $v_m = P(v_{m-1})$, with some user-supplied initial guess $v_0$.
Given such a method, the associated \ac{PSIM} is obtained by pushing the prior $\eta_0$ through the map $P^m$ defined by composing $P$ with itself $m$ times.
We will restrict attention to affine $P$, i.e.\ $P(v) = \itermat v + \itervec$, where $\itermat \in \reals^{d \times d}$ while $\itervec \in \reals^d$.
Then, for Gaussian $\eta_0 = \mathcal{N}(0, \gramian^{-1})$ the output of the \ac{PSIM} is given by
\begin{subequations}
    \begin{align}
        \eta_m &= \mathcal{N}(v_m, C_m) \\
        v_m &= P^m(0) = \sum_{i=0}^{m-1} \itermat^i \itervec \\
        C_m &= \itermat^m \gramian^{-1} (\itermat^\top)^m. \label{eq:psim_cov}
    \end{align}
\end{subequations}

Since the output of a \ac{PSIM} is not a Bayesian posterior, several important questions arise: (i) when does the posterior contract around the truth, and (ii) is the output calibrated in the sense of \cref{def:calib}.
For (i), \citet[Proposition 2]{Cockayne2021PIMs} establishes that provided the underlying stationary iterative method converges to the true solution\footnote{This is not guaranteed; they converge to the truth only when the spectral radius of $\itermat$ is below 1 and if they are \emph{completely consistent} (see \cref{sec:LSIM}).}, the \ac{PSIM} contracts around the true solution, and does so at the same rate as the error converges in any norm on $\reals^d$.
For (ii), \citet[Propositions 7 and 10]{Cockayne2021PIMs} show that any \ac{PSIM} based on appropriate affine $P$ is strongly calibrated in the sense of \cref{def:calib}, provided $\itermat$ is diagonalisable (over $\mathbb{C}$).
This is a fairly mild restriction considering the density of diagonalisable matrices \citep[Chapter~2]{GorodentsevAlgebraII}.

In the next sections we will demonstrate how \acp{PSIM} can be adapted to work with \acp{CAGP}, and discuss transfer of calibratedness in this setting.

\section{CALIBRATED COMPUTATION-AWARE GPs} \label{sec:calib_cagp}
In this section we prove that \acp{CAGP} are calibrated if a calibrated \acp{PLS} is used to solve \cref{eq:the_system}.
First we observe that the reparameterisation in \cref{eq:gp_posterior} can be formulated as Bayesian inference with an alternative observation model.

\begin{proposition} \label{prop:reparam_obs_model}
    It holds that $f \mid y$ from \cref{eq:gp_posterior} is equal to the posterior $f \mid v$ from Bayesian inference under the observation model:
    $$
    v \mid f = G^{-1} b = \gramian^{-1} (f(X) - m_0(X)) + \bar{\zeta}
    $$
    where $\bar{\zeta} \sim \mathcal{N}(0, \sigma^2 \gramian^{-2})$.
\end{proposition}

The next corollary establishes that under the observation model used in \cref{prop:reparam_obs_model}, the prior distribution adopted in \acp{CAGP} is correct in a subjective Bayesian sense.

\begin{corollary} \label{corr:cagp_req_prior}
    The a-priori marginal distribution of $v$ is $v \sim \mathcal{N}(0, \gramian^{-1})$.
\end{corollary}

This is an important result for calibration, since to be able to talk about calibrated posteriors for \cref{eq:the_system} we first need to know that if the prior on $f$ is correct, the prior on $v$ is also correct.

The next result is the central result of the paper, establishing calibratedness of \acp{CAGP} when a calibrated \ac{PLS} is used.

\begin{theorem} \label{thm:itergp_calib}
    Suppose that $f \sim \mathcal{GP}(m_0, k)$ and $y \mid f = f(X) + \zeta$, where $k$ is a positive definite kernel.
    Further, suppose that the PLS in \cref{eq:cagp_req_posterior} is a learning procedure that is calibrated for $\mathcal{N}(0, \gramian^{-1})$, and satisfies the following conditions:
    \begin{enumerate}
        \item $D$ is independent of $y$. \label{cond:indep}
        \item $\textup{Cov}(f(X'), \bar{v}) = K(X', X) \gramian^{-1} \textup{Cov}(y, \bar{v})$. \label{cond:cross}
    \end{enumerate}
    Then the \ac{CAGP} posterior is calibrated for the original \ac{GP} prior and the original data generating model.
\end{theorem}

Regarding the conditions above, as mentioned previously, Condition \ref{cond:indep} is satisfied by most \acp{PLS}, with the notable exception of those where some calibration procedure has been applied to choose the prior.
Condition \ref{cond:cross} seems technical, but is in fact satisfied under mild conditions.

\begin{corollary} \label{corr:cross}
    Suppose that the assumptions of \cref{thm:itergp_calib} are satisfied, and further that $\bar{v}$ is an affine map of $y$, i.e.\ $\bar{v} = M y + g$.
    Then the \ac{CAGP} posterior is calibrated for the original \ac{GP} prior and original data generating model.
\end{corollary}

Note that the requirements of \cref{corr:cross} are satisfied by any Bayesian procedure, as well as for the probabilistic iterative methods we will introduce in the next section.
An important exception is \ac{CAGP}-CG since, as mentioned in \cref{sec:BPLS}, in this case the map is not affine owing to dependence of $S_m$ on $v^\star$.
Having established this result, we next proceed to show how \acp{PSIM} can be integrated with \acp{CAGP} to provide calibrated uncertainty.

\section{PROBABILISTIC STATIONARY ITERATIVE METHODS WITH THE INVERSE PRIOR} \label{sec:inverse_pims}

To embed a probabilistic stationary iterative method in a \ac{CAGP} we need to obtain an output measure as in \cref{eq:cagp_req_posterior}.
We now show that for all reasonable \acp{PSIM}, $C_m$ from \cref{eq:psim_cov} can be rearranged to have this structure.

\subsection{Convergent Linear Stationary Iterative Methods}\label{sec:LSIM}
As mentioned in \cref{sec:PIMs}, convergence of stationary iterative methods is not guaranteed for all affine maps $P$.
We therefore limit attention to \emph{completely consistent} methods \citep[Section 3.2 and 3.5]{Young1971}.
These methods have the property that, if they converge, they are guaranteed to converge to the true solution, and therefore they are the widest class of reasonable stationary iterative methods to use for solving a linear system.

For nonsingular $\gramian$, any completely consistent iterative method can be written in the form
$\itermat = I - \ccitermat \gramian, \; \itervec = \ccitermat b$
where $\ccitermat$ is a nonsingular matrix.
The next proposition shows that the iterations of any \ac{PSIM} based on a completely consistent iterative method can be written in a form commensurate with \acp{CAGP}.

\begin{proposition} \label{prop:pim_itergp}
    Let $\eta_m$ be the $m$\textsuperscript{th} iterate of a probabilistic iterative method whose underlying stationary iterative method is completely consistent.
    Then $\eta_m \sim \mathcal{N}(\bar{v}_m, \gramian^{-1} - D_m)$, where $\bar{v}_m$ is the $m$\textsuperscript{th} iterate of the stationary iterative method with $v_0 = 0$, while $D_0 = 0$ and
    \begin{equation}
        D_{m} = \ccitermat + \ccitermat^\top - \ccitermat\gramian\ccitermat^\top + (I - \ccitermat\gramian) D_{m-1} (I - \ccitermat\gramian)^\top.
    \end{equation}
\end{proposition}

While it is useful to know that this holds for any completely consistent probabilistic iterative method, it is not clear that the above computations can be made to be efficient.
Moreover, we still need to ensure that the iterative method converges, as it will be highly problematic for embedding within \ac{GP} regression otherwise.
In the next section we consider a particular instance of a probabilistic iterative method which is provably convergent for any symmetric positive definite matrix.

\subsection{Gauss-Seidel} \label{sec:GaussSeidal}

The Gauss-Seidel method partitions $\gramian = L + U$, where $L$ is the lower-triangular part of $\gramian$ and $U$ is the strict upper triangular part.
We then take $\itermat = -L^{-1} U$ and $\itervec = L^{-1}b$.
\citet[Theorem 11.2.3]{GVL} establishes that Gauss-Seidel converges to the true solution for any initial guess $v_0$ provided $\gramian$ is symmetric positive-definite, making it particularly attractive for \acp{CAGP}.

To apply \cref{prop:pim_itergp} we must first identify $\ccitermat$.
Note that $U = \gramian - L$, so that $\itermat = -L^{-1} U = I - L^{-1} \gramian$; thus $\ccitermat = L^{-1}$.
Also note that if $\gramdiag$ is the diagonal of $\gramian$, since $\gramian$ is symmetric positive definite, we have that $U = (L - \gramdiag)^\top$.
The $D_i$ from \cref{prop:pim_itergp} can then be simplified:
\begin{align*}
    L^{-1} \gramian L^{-\top} &= L^{-1} (L + L^\top - \gramdiag) L^{-\top} \\
    &= L^{-1} + L^{-\top} -  L^{-1} \gramdiag L^{- \top}
\end{align*}
so that $D_1 = L^{-1} \gramdiag L^{-\top}$, and
\begin{align*}
    D_i &= L^{-1} \gramdiag L^{-\top} + \itermat D_{i-1} \itermat^\top \\
    &= L^{-1} \gramdiag L^{-\top} + L^{-1} U D_{i-1} U^\top L^{-\top} .
\end{align*}
We therefore have the following non-recursive expression for the downdate:
\begin{equation}
    D_m = \sum_{i=0}^{m-1} (L^{-1} U)^i L^{-1} \gramdiag L^{-\top} (U^\top L^{-\top})^i.
\end{equation}

We can also establish several important properties of this \ac{PLS}, in the following proposition:

\begin{proposition} \label{prop:gs_properties}
    The covariance matrix $G^{-1} - D_m$ from probabilistic Gauss-Seidel has rank $d-1$ for all $m \geq 1$, and its null space is equal to $\textup{span}(L^{\top} e_d)$, where $e_d$ is the $d$\textsuperscript{th} Euclidean basis vector.
\end{proposition}

Considering \cref{sec:itergp} we can therefore implement \textsc{cagp\_pls} with Gauss-Seidel as described in \cref{alg:downdate_gs}.

\begin{algorithm}
\caption{\ac{CAGP}-GS} \label{alg:downdate_gs}
\begin{algorithmic}[1]
\Function{\textsc{cagp\_pls\_gs}}{$G$, $b$, $V$, $m$}
\State $Z_1 = L^{-\top} V^\top$
\State $z = L^{-1} b$
\State $v_1 = z$
\State $\tilde{Z}_1 = \gramdiag^\frac{1}{2} Z_1$
\For{$i = 2$ to $m$}
    \State $v_i = z - L^{-1} U v_{i-1}$
    \State $Z_i = L^{-\top} U^\top Z_{i-1}$
    \State $\tilde{Z}_i = \gramdiag^\frac{1}{2} Z_i$
\EndFor
\State \Return $\tilde{v} = V v_i$, $\tilde{D} = \sum_{i=1}^m \tilde{Z}_i^\top \tilde{Z}_i$
\EndFunction
\end{algorithmic}
\end{algorithm}

\subsubsection{Complexity} \label{sec:complexity}

Since $L^{-1}$ is lower triangular, the action of $L^{-1}$ and $L^{-\top}$ can be computed using forward and back substitution, having complexity $\mathcal{O}(d^2)$, where $d$ is the size of the training data.
With for $m$ iterations and $d_\textsc{test}$ testing points, the computational complexity of \cref{alg:downdate_gs} is thus $\mathcal{O}(m d d_\textsc{test}(d + d_\textsc{test}))$.

As mentioned in \cref{sec:BPLS}, the cost of \ac{CAGP}-CG is $\mathcal{O}(m d^2)$ to compute the posterior over $v$.
Once this has been computed, computing the implied $\tilde{v}$ and $\tilde{D}$ costs, respectively $\mathcal{O}(d d_\textsc{test})$ and $\mathcal{O}(m d_\textsc{test}  (d + d_\textsc{test}))$, for an overall complexity of $\mathcal{O}(m(d^2 + dd_\textsc{test} + d_\textsc{test}^2))$.

In terms of memory, for \ac{CAGP}-GS we need to store the matrix $Z_i$ for complexity $\mathcal{O}(dd_\textsc{test})$; computed factors can be saved to disk and loaded later to compute $\tilde{D}$ (which is only $d_\textsc{test} \times d_\textsc{test}$).
For \ac{CAGP}-CG only $\mathcal{O}(d)$ memory is required at execution time.
This ignores storage of $G$ (required for both algorithms); however since each algorithm requires only the action of $G$ (or $L$, $U$) on matrices / vectors, a matrix-free implementation is possible (though this is not explored in this paper).
Also note that this highlights that to apply \ac{CAGP}-CG on a new set of test points does not require rerunning the algorithm, while for \ac{CAGP}-GS we need to do so.

Clearly \ac{CAGP}-GS has a higher complexity than \ac{CAGP}-CG, though under the assumption that $d \gg m, d_\textsc{test}$ the leading order of both algorithms is $\mathcal{O}(d^2)$, so in this setting the costs should still be comparable.
In the next section we will consider the empirical performance of \ac{CAGP}-GS compared to \ac{CAGP}-CG.

\section{EXPERIMENTS} \label{sec:simulations}

A Python implementation of \cref{alg:downdate_gs} is available at \texttt{https://github.com/hegdedisha/CalibratedCAGP}.

\subsection{Synthetic Problem} \label{sec:synthetic}

We first consider a synthetic test problem, so that we can test for calibratedness.
We take $m_0 = 0$ and $k$ to be a Matèrn $\sfrac{3}{2}$ covariance with amplitude set to $1$, and will vary the length-scale.
For the data-generating model we set $\sigma = 0.1$.
We set our domain to be $D = [0,1]^2$ and generate our training points by sampling $400$ points uniformly at random.
Test points are a regular grid with spacing $0.05$, i.e.\ generating $21 \times 21$ equally spaced points for a total of $d = 441$ points. The underlying true function is taken to be sample from the prior, so that the calibration guarantee from \cref{corr:cross} can be tested.

\begin{figure}
\includegraphics[width=0.5\textwidth]{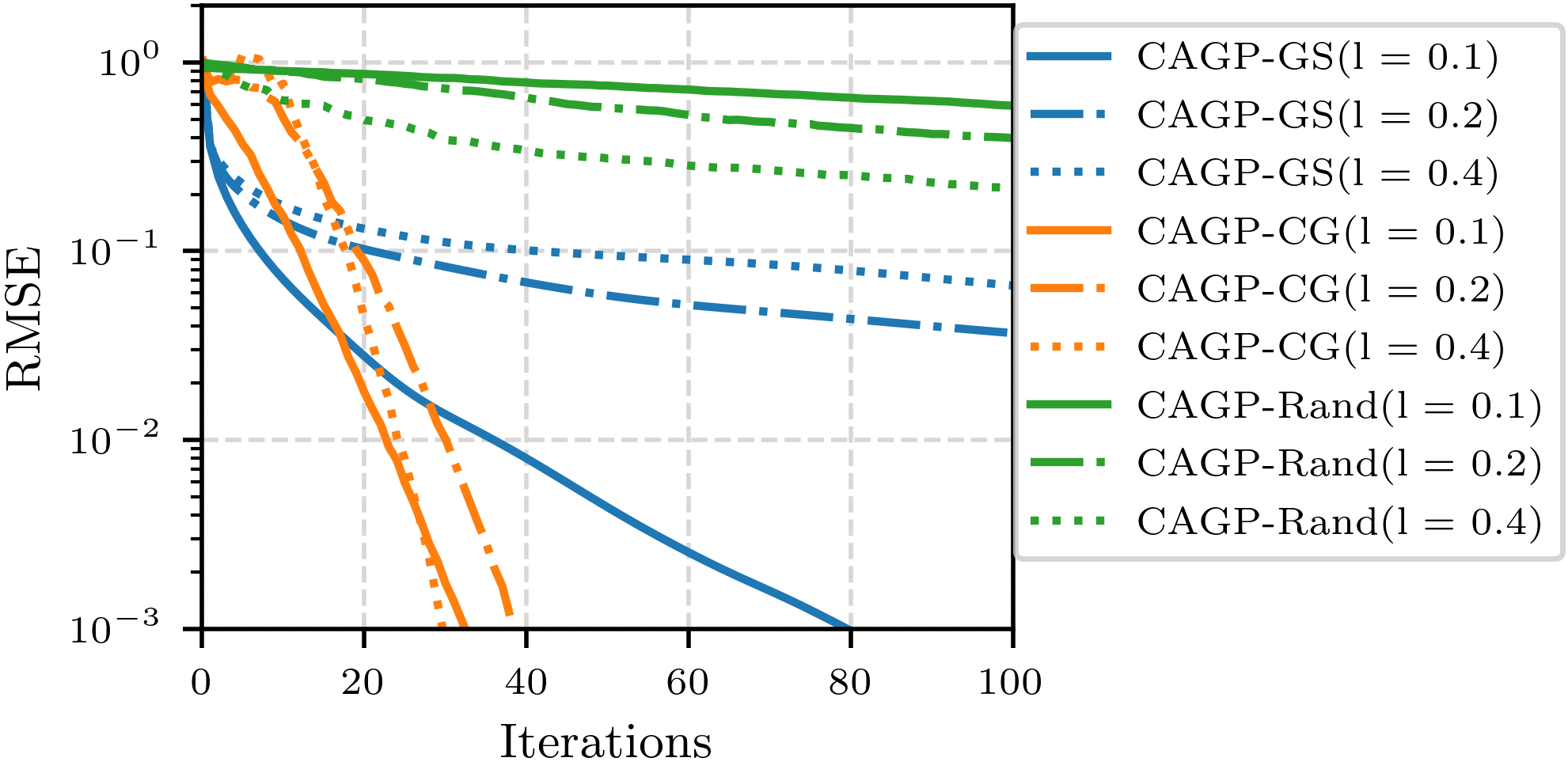}
\caption{RMSE for the synthetic problem in \cref{sec:synthetic}.} \label{fig:synthetic:rmse_vary_l}
\end{figure}

Plots of \ac{RMSE} of the posterior mean as a function of $m$, with length scales set to $0.1, 0.2$ and $0.4$, averaged over 50 runs, can be seen in \cref{fig:synthetic:rmse_vary_l}.
We compare \ac{CAGP}-GS to both \ac{CAGP}-CG and \ac{CAGP}-Rand, a method that uses a Bayesian \ac{PLS} with $S_i$ having IID normal entries.
This approach should be calibrated, but shows slow convergence typical of most Bayesian \acp{PLS} not using \ac{CG} directions.
Unexpectedly, the posterior mean for \ac{CAGP}-GS initially converges \emph{faster} than that from \ac{CAGP}-CG for all length-scales, though ultimately it is overtaken by \ac{CG}.
However, in expensive problems where few iterations can be performed, this suggests that \ac{CAGP}-GS should be preferred to \ac{CAGP}-CG owing to its initially faster convergence with calibration guarantees.
As a function of the length-scale, it appears that this behaviour is less pronounced for smaller values, in which cases the matrix will typically be better conditioned.

\begin{figure}
\includegraphics[width=0.5\textwidth]{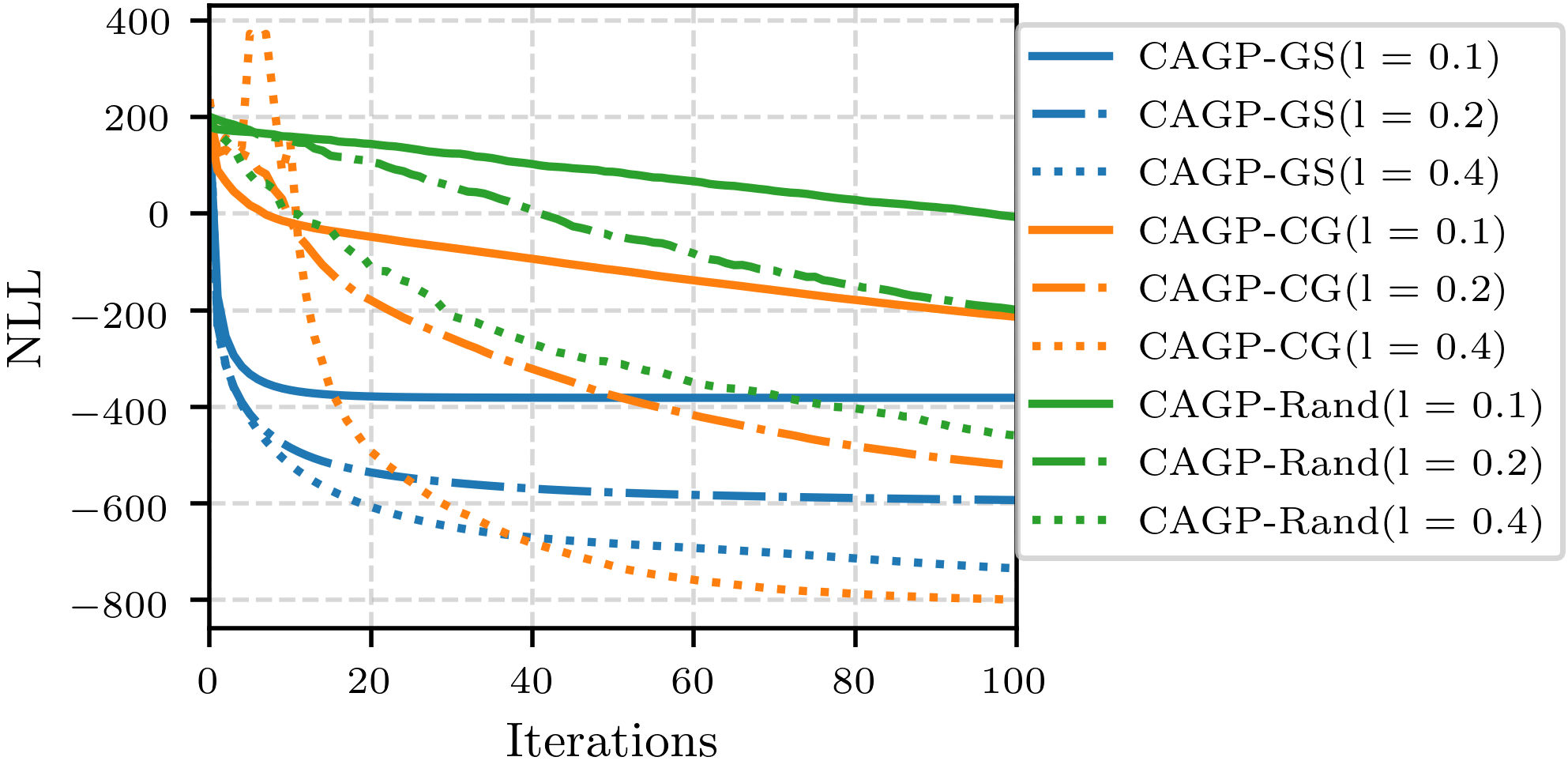}
\caption{NLL for the synthetic problem in \cref{sec:synthetic}.} \label{fig:synthetic:nll_vary_l}
\end{figure}

The \ac{NLL} plot for the same problem in \cref{fig:synthetic:nll_vary_l} shows the added advantage of calibration for \ac{CAGP}-GS. \ac{CAGP}-GS outperforms \ac{CAGP}-CG in \ac{NLL} for more number of iterations than it does in \ac{RMSE}, due to better calibrated posterior.

\begin{figure*}
    \begin{subfigure}{0.32\textwidth}
        \includegraphics[width=\textwidth]{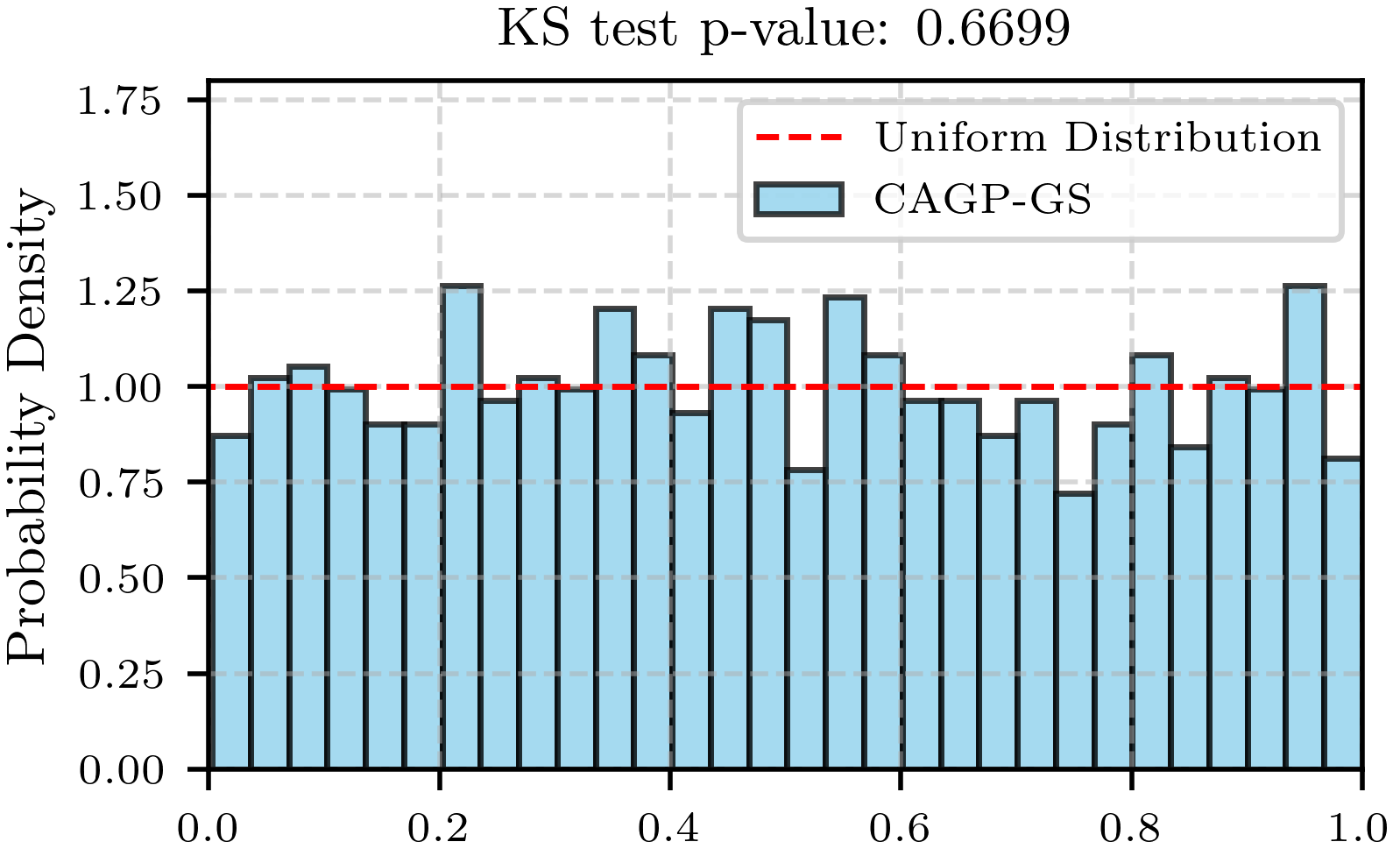}
        \caption{CAGP-GS} \label{fig:synthetic:sbc-gs}
    \end{subfigure}
    \begin{subfigure}{0.32\textwidth}
        \includegraphics[width=\textwidth]{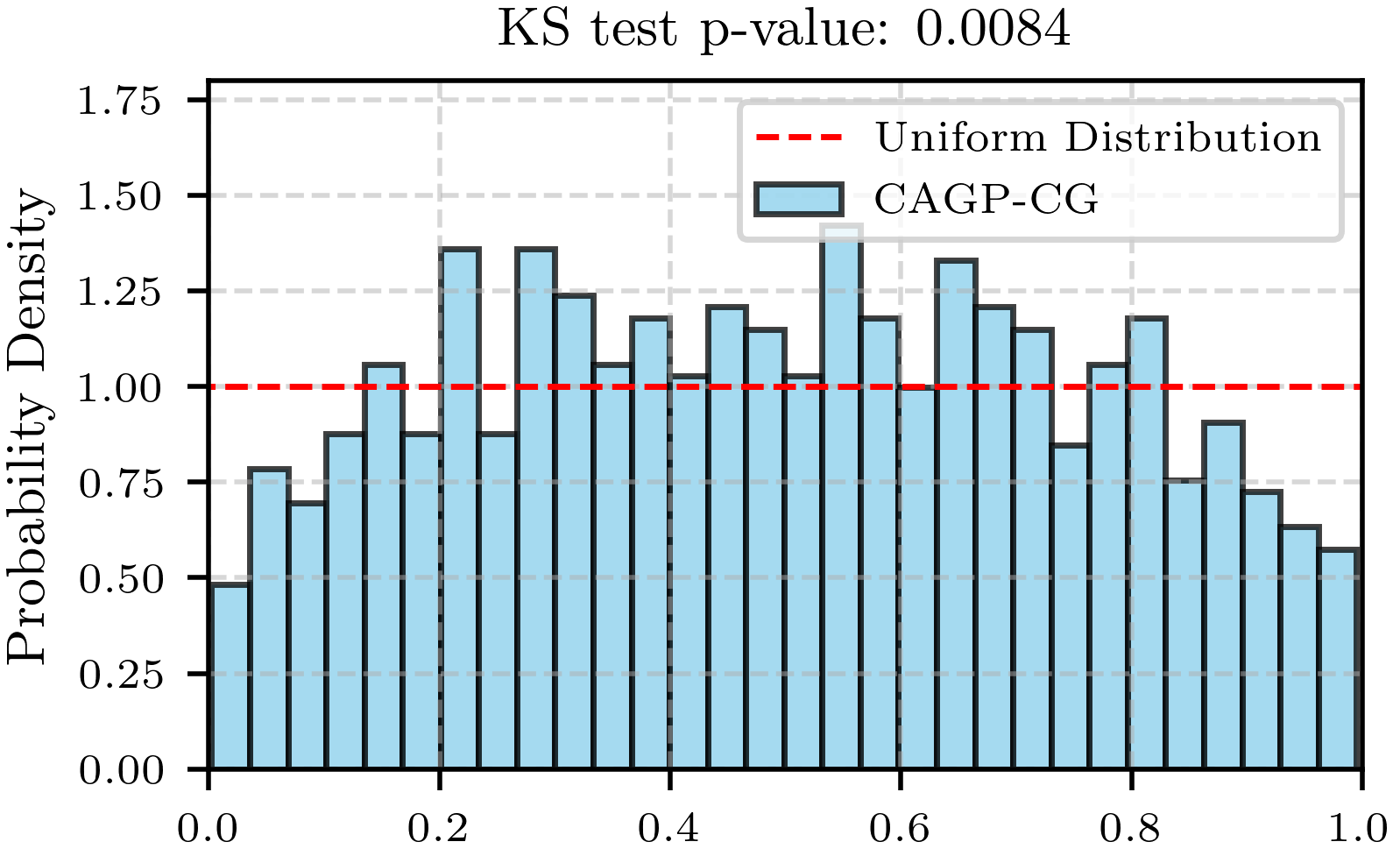}
        \caption{CAGP-CG} \label{fig:synthetic:sbc-cg}
    \end{subfigure}
    \begin{subfigure}{0.32\textwidth}
        \includegraphics[width=\textwidth]{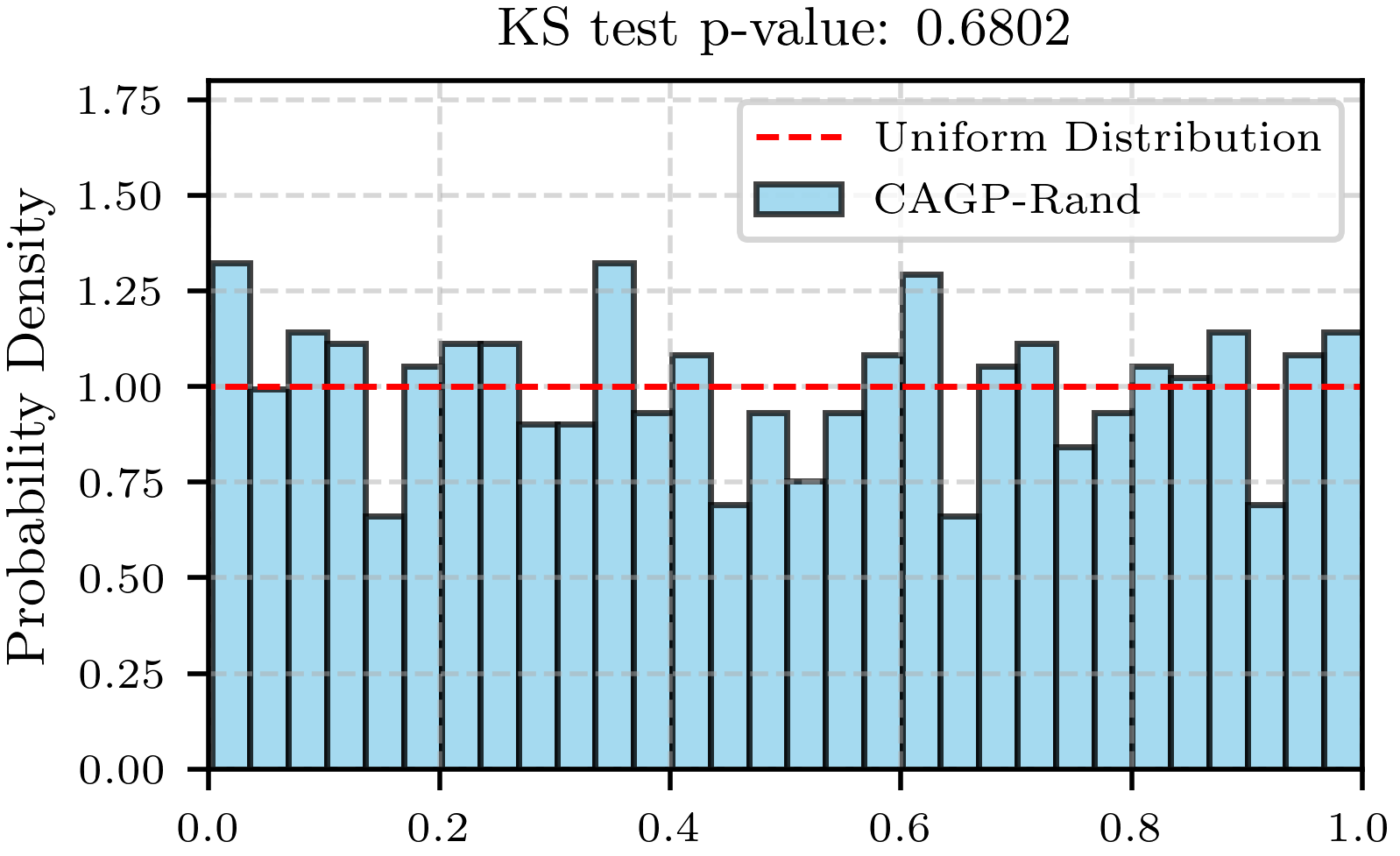}
        \caption{CAGP-Rand} \label{fig:synthetic:sbc-rand}
    \end{subfigure}
    \caption{Simulation-based calibration results for the synthetic problem described in \cref{sec:synthetic} after $m = 5$ iterations.} \label{fig:synthetic:sbc}
    \end{figure*}

In \cref{fig:synthetic:sbc} we highlight the uncertainty quantification properties of \ac{CAGP}-GS compared to \ac{CAGP}-CG and \ac{CAGP}-Rand using the simulation-based calibration method described in \cref{alg:sbc} for $N_\textsc{sim}=1000$ simulations, with length-scale now fixed to $\ell=0.2$ and $m=5$ iterations.
We also report the result of a Kolmogorov-Smirnov test for uniformity.
As implied by \cref{thm:itergp_calib}, both \ac{CAGP}-GS and \ac{CAGP}-Rand are calibrated ($p$-values 0.6689 and 0.6802), while \ac{CAGP}-CG shows the expected inverted U-shape characteristic of an overly conservative posterior and has a $p$-value of 0.0084.

\begin{figure*}[h!]
    \includegraphics[width=\textwidth]{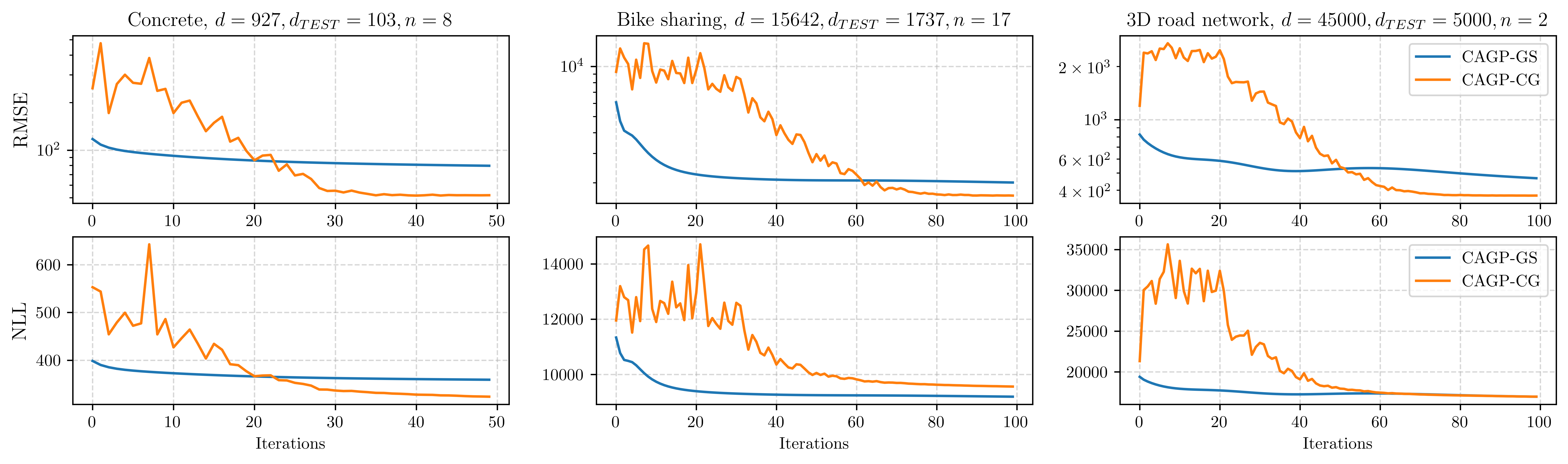}
    \caption{RMSE and NLL for different datasets in \cref{sec:benchmark}}
    \label{fig:benchmark:rmsenll} 
\end{figure*}
\subsection{UCI Benchmark Datasets} \label{sec:benchmark}

In this section we compare the performance of \ac{CAGP}-GS and \ac{CAGP}-CG on three datasets from the UCI Machine Learning Repository: Concrete Compressive Strength dataset \citep{concrete_compressive_strength_165}, Bike Sharing dataset \citep{bike_sharing_275} and 3D Road Network dataset \citep{3d_road_network}.
These have varying number of explanatory variables and varying size of the dataset

As seen in the previous results, \cref{fig:benchmark:rmsenll} shows that \ac{CAGP}-GS outperforms \ac{CAGP}-CG in \ac{RMSE} for initial iterations. And due to better calibration, \ac{CAGP}-GS also outperforms \ac{CAGP}-CG in \ac{NLL} for higher iteration numbers the last two datasets (bike sharing and 3D road network). 
Additional improvement in terms of \ac{NLL} is not seen for the first dataset, and we speculate that this could be because of a small number of data points. Since we not ``in-model'' as in \cref{sec:synthetic}, the small data size could possibly lead to model mismatch and, thus, suboptimal performance in \ac{NLL}. However, there is still a sizeable region where \ac{CAGP}-GS outperforms \ac{CAGP}-CG in \ac{NLL}.

\subsection{ERA5 Regression} \label{sec:era5}

\begin{figure*}
    \includegraphics[width=\textwidth]{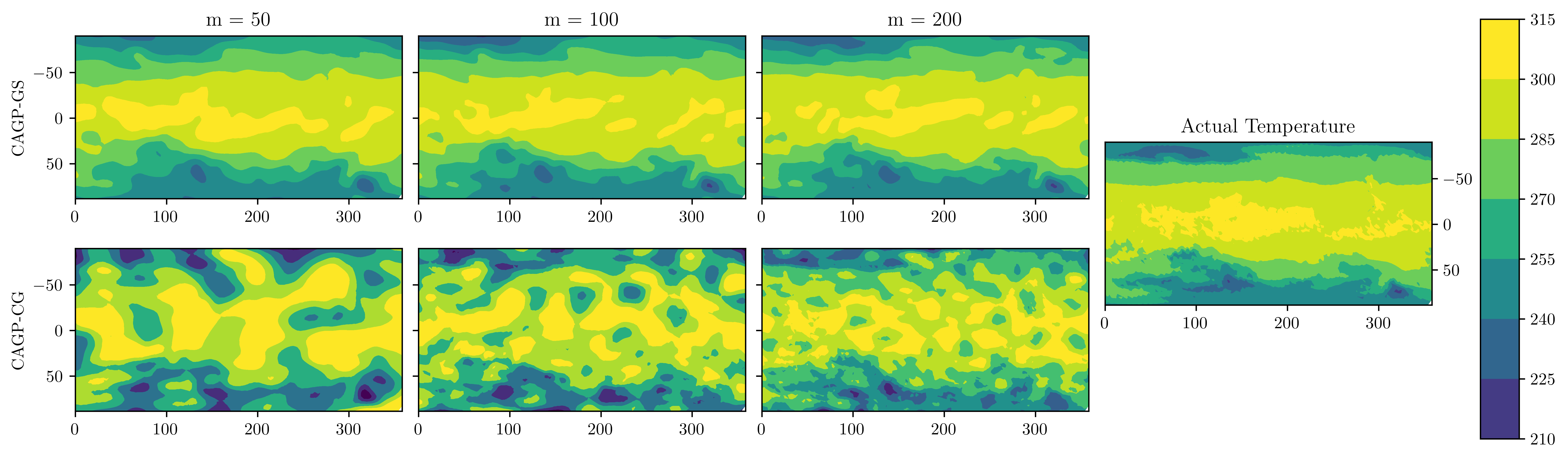}
    \caption{Sequences of posterior means for the large-scale regression from \cref{sec:era5}, as a function of $m$. The x and y axes represent longitudes and latitudes respectively, and the contours indicate the temperature in Kelvin.}
    \label{fig:era5:mean} 
\end{figure*}

In this section we solve a geospatial regression problem on the ERA5 global 2 metre temperature dataset \citep{era5_2023}.
This is a reanalysis dataset with approximately 31 km resolution, and has temporal coverage from 1940 to present. For the purposes of this demonstration we limit our attention to a single timestamp on 1st January 2024 at 00.00. This results in a grid of a total of $\approx 1$ million points.
To obtain matrices that can be more easily represented in memory these points were downsampled as described in the results below.

To fix a prior we used a Matèrn $\sfrac{3}{2}$ covariance function, and optimised hyper-parameters by maximising marginal likelihood (see \citet[Sec.~2.2]{rasmussen_gaussian_2005}) on a coarse uniform grid of $72 \times 144$ points on the globe.
We use a constant prior mean fixed to the average of the data points.

We first examine qualitative convergence of the method by presenting plots of the posterior mean obtained from \ac{CAGP}-GS and \ac{CAGP}-CG for a uniformly spaced grid of $28,800$ training and $7,225$ test points in  \cref{fig:era5:mean}.
Interestingly the \ac{CAGP}-GS posterior means are considerably smoother than the \ac{CAGP}-CG posterior means; this smoother convergence may be more desirable in low iteration number regimes, and is likely due to known smoothing properties of Gauss-Seidel (see e.g.\ \citet[Section 5.5]{xu_algebraic_2017}).

In \cref{fig:era5:timings}, we compare the wall-time taken to compute the posterior mean and covariance using \ac{CAGP}-GS and \ac{CAGP}-CG. In each plot we vary one parameter, fixing the others to $m = 80, d = 10638, d_\textsc{test} = 25$.
Results were computed on a high-performance computing service on a single node with 40 cores.
The results are mostly as expected from \cref{sec:complexity}.
With increasing iterations, \ac{CAGP}-GS and \ac{CAGP}-CG scale similarly, with \ac{CAGP}-GS performing slightly better for higher iterations.
With increasing $d$ \ac{CAGP}-CG is initially faster, but \ac{CAGP}-GS takes over in the later part.
For small $d_\textsc{test}$ \ac{CAGP}-GS performs better, but \ac{CAGP}-CG improves as $d_\textsc{test}$ increases and becomes comparable to $d$.

Finally, we test calibratedness for a uniformly spaced grid of $d=64,700$ training points and $d_\textsc{test} = 100$ held out test points.
Note that in this setting \cref{thm:itergp_calib} does not provide a calibration guarantee, and in particular our theory says nothing about calibratedness on held-out data.
Nevertheless, it is interesting to see whether some version of calibratedness is obtained in a practical setting.
For this $d$, $G$ requires around 30 GB of storage, so is around the largest that can practically be considered without matrix-free implementations.
Histograms of $t_i = \Phi\left(\frac{\tilde{m}(x_i) - f(x_i)}{sd(x_i)}\right)$ for test points $x_i$ are shown in \cref{fig:era5:zscores}.
The Kolmogorov-Smirnov test for uniformity gives a $p$-value of $0.4707$ for \ac{CAGP}-GS and $5.0112 \times 10^{-1}$ \ac{CAGP}-CG.
This shows that \ac{CAGP}-GS is closer to calibrated, and \ac{CAGP}-CG remains miscalibrated.

\section{CONCLUSION} \label{sec:conclusion}

The theoretical and computational results presented above provide a clear motivation for considering \acp{PLS} other than BayesCG for \acp{CAGP}. 
Mean convergence is faster for small $m$ as shown in \cref{fig:synthetic:rmse_vary_l}, and the posterior mean plots in \cref{fig:era5:mean} show recoveries that, we would argue, retain more of the smoothness and structure of the prior than those for \ac{CAGP}-CG.
Further, with an appropriate \ac{GP} calibration approach it appears we can obtain reasonably well-calibrated \acp{CAGP} even outside of the synthetic calibration guarantees of \cref{thm:itergp_calib}. 

The principle downsides of \ac{CAGP}-GS are (i) higher computational complexity and (ii) worse scaling for large $m$.
On (i) we would argue that this approach should only be used in a small $d_\textsc{test}$ regime which, in very large scale regression problems, is likely to be a limitation in any case.
For (ii) we would argue that small $m$ is the regime in which well-calibrated \acp{CAGP} are most attractive, since for larger $m$ the added computational uncertainty is dominated by the mathematical uncertainty.
Moreover, we feel that the calibration benefits are enough to justify these disadvantages.

There are several interesting future research directions.
First, we have not explored matrix-free methods to scale to ``big data'' problems.
Efficient, parallelisable matrix-free implementations of $L^{-\top}$ are challenging due to the inherently sequential nature of backward substitution, but would nevertheless allow further scaling.
Second, we would like to explore further accelerations of calibrated \acp{CAGP}.
One avenue that seems promising is to combine \ac{CAGP}-GS and \ac{CAGP}-CG; this would sacrifice calibratedness but, if GS is either used in the initial convergence period or interleaved with CG iterations to provide smoothing, this may result in superior mean convergence.

\subsubsection*{Acknowledgements}
JC was funded in part by EPSRC grant EP/Y001028/1. The authors thank Marvin Pförtner, Jonathan Wenger and Dave Woods for helpful discussion.
\bibliographystyle{apalike}
\bibliography{citations}

\clearpage
\section*{Checklist}

 \begin{enumerate}

 \item For all models and algorithms presented, check if you include:
 \begin{enumerate}
   \item A clear description of the mathematical setting, assumptions, algorithm, and/or model. [Yes]
   \item An analysis of the properties and complexity (time, space, sample size) of any algorithm. [Yes]
   \item (Optional) Anonymized source code, with specification of all dependencies, including external libraries. [No] Code will be made available for the camera-ready submission
 \end{enumerate}

 \item For any theoretical claim, check if you include:
 \begin{enumerate}
   \item Statements of the full set of assumptions of all theoretical results. [Yes]
   \item Complete proofs of all theoretical results. [Yes] All the proofs can be found in the Supplementary materials.
   \item Clear explanations of any assumptions. [Yes]
 \end{enumerate}

 \item For all figures and tables that present empirical results, check if you include:
 \begin{enumerate}
   \item The code, data, and instructions needed to reproduce the main experimental results (either in the supplemental material or as a URL). [No] Source of data is provided and code will be made available for the camera-ready submission
   \item All the training details (e.g., data splits, hyperparameters, how they were chosen). [Yes]
         \item A clear definition of the specific measure or statistics and error bars (e.g., with respect to the random seed after running experiments multiple times). [Yes] 
         \item A description of the computing infrastructure used. (e.g., type of GPUs, internal cluster, or cloud provider). [Yes] 
 \end{enumerate}

 \item If you are using existing assets (e.g., code, data, models) or curating/releasing new assets, check if you include:
 \begin{enumerate}
   \item Citations of the creator If your work uses existing assets. [Yes] Existing data is used in \cref{sec:era5} and creator is cited.
   \item The license information of the assets, if applicable. [Yes] 
   \item New assets either in the supplemental material or as a URL, if applicable. [Not Applicable]
   \item Information about consent from data providers/curators. [Yes]
   \item Discussion of sensible content if applicable, e.g., personally identifiable information or offensive content. [Not Applicable]
 \end{enumerate}

 \item If you used crowdsourcing or conducted research with human subjects, check if you include:
 \begin{enumerate}
   \item The full text of instructions given to participants and screenshots. [Not Applicable]
   \item Descriptions of potential participant risks, with links to Institutional Review Board (IRB) approvals if applicable. [Not Applicable]
   \item The estimated hourly wage paid to participants and the total amount spent on participant compensation. [Not Applicable]
 \end{enumerate}

 \end{enumerate}

 \appendix

\numberwithin{figure}{section}
\numberwithin{table}{section}
\numberwithin{algorithm}{section}

\numberwithin{proposition}{section}
\numberwithin{equation}{section}

\onecolumn
\thispagestyle{empty}
\aistatstitle{Calibrated Computation-Aware Gaussian Processes: Supplementary Materials} 

\section{Proofs of Theoretical Results}

\begin{proof}[Proof of \cref{prop:reparam_obs_model}]
    This is demonstrated by direct calculation.
    We have that $f(X')$ and $v$ are jointly Gaussian:
    \begin{align*}
        \begin{bmatrix}
            f(X') \\ v
        \end{bmatrix}
        &\sim
        \mathcal{N}\left(
            \begin{bmatrix}
                m_0(X') \\
                0
            \end{bmatrix}
            ,
            \begin{bmatrix}
                k(X', X') & k(X', X) \gramian^{-1} \\
                \gramian^{-1} k(X, X') & \gramian^{-1} k(X, X) \gramian^{-1} + \sigma^2\gramian^{-2}
            \end{bmatrix}
        \right).
    \end{align*}
    Clearly
    \begin{align*}
        \gramian^{-1} k(X, X) \gramian^{-1} + \sigma^2 \gramian^{-2} &= \gramian^{-1} (k(X, X) + \sigma^2 I) \gramian^{-1} \\
        &= \gramian^{-1} 
    \end{align*}
    and so
    \begin{align*}
        \begin{bmatrix}
            f(X') \\ v
        \end{bmatrix}
        &\sim
        \mathcal{N}\left(
            \begin{bmatrix}
                m_0(X') \\
                0
            \end{bmatrix}
            ,
            \begin{bmatrix}
                k(X', X') & k(X', X) \gramian^{-1} \\
                \gramian^{-1} k(X, X') & \gramian^{-1} 
            \end{bmatrix}
        \right).
    \end{align*}
    Applying the Gaussian conditioning formula we obtain
    \begin{align*}
        \mathbb{E}[f(X') \mid v] &= m_0(X') + k(X', X) \gramian^{-1} \gramian v \\
        &= m_0(X') + k(X', X)v \\
        \mathbb{V}[ f(X') \mid v] &= k(X', X') - k(X', X) \gramian^{-1} \gramian \gramian^{-1} k(X, X') \\
        &= k(X', X') - k(X', X) \gramian^{-1}  k(X, X')
    \end{align*} 
    as required.
\end{proof}

\begin{proof}[Proof of \cref{corr:cagp_req_prior}]
    This can be verified by inspection of the joint distributions in the proof of \cref{prop:reparam_obs_model}.
\end{proof}


\begin{proof}[Proof of \cref{thm:itergp_calib}]
    To prove this we use the definition of calibratedness from \cref{def:calib}.
    First let $\tilde{C} = \tilde{k}(X', X')$ and $\bar{C} = \bar{k}(X', X')$.
    We will similarly abbreviate $\tilde{m} = \tilde{m}(X')$, $\bar{m} = \bar{m}(X')$ and $f = f(X')$.

    We first establish that $\tilde{C}$ is full rank; this is trivial since $\tilde{C} = \bar{C} + k(X', X) C_m k(X, X')$, where the latter term is positive semidefinite owing to positive semidefiniteness of $C_m$.
    Thus, $\tilde{C} \succeq \bar{C}$, and since $\bar{C}$ is positive definite, $\tilde{C}$ is full rank.
    For the purposes of checking calibration of the GP, we therefore do not need to worry about range and null spaces of $\tilde{C}$, so the quantity of interest is:
    \begin{align}
        \tilde{C}^{-\frac{1}{2}} (\tilde{m} - f) &= \tilde{C}^{\frac{1}{2}} \tilde{C}^{-1} (\tilde{m} - f) \label{eq:calib_req}\\
        &= \tilde{C}^{\frac{1}{2}} \tilde{C}^{-1} (\bar{m} - f) + \tilde{C}^{\frac{1}{2}} \tilde{C}^{-1} k(X', X)(\bar{v} - v) \label{eq:calib_split}
    \end{align}

    \paragraph{Preliminary Transformations.}
    We start by applying the matrix inversion lemma to obtain a more useful expression for $\tilde{C}$.
    Since $C_m = G^{-1} - D_m$ is not assumed to be full rank we let $R$, $N$ be bases of its row and null spaces of $C_m$ respectively, and such that $U = \begin{bmatrix} R & N \end{bmatrix}$ is unitary.
    We then have that
    \begin{align*}
        \tilde{C} &= \bar{C} + k(X', X) C_m k(X, X') \\
        &= \bar{C} + k(X', X) UU^\top  C_m UU^\top k(X, X') \\
        &= \bar{C} + k(X', X) R C_m^R R^\top k(X, X')
    \end{align*}
    where $C_m^R = R^\top C_m R$, since $C_m N v = 0$ by definition for all $v$.
    Applying the matrix inversion lemma we get
    \begin{align*}
        \tilde{C}^{-1} &= \bar{C}^{-1} - \bar{C}^{-1} k(X', X) \Sigma^R k(X, X') \bar{C}^{-1} \\
        &= (I-P)\bar{C}^{-1} 
    \intertext{where}
        P &= \bar{C}^{-1} k(X', X) \Sigma^R k(X, X') \\
        \Sigma^R &= R \Sigma^{-1} R^\top \\
        \Sigma &= (C_m^R)^{-1}  +R^\top  k(X, X') \bar{C}^{-1} k(X', X) R 
    \end{align*}
    We also have
    \begin{align*}
        \tilde{C}^{-1} k(X', X) &= \left[\bar{C}^{-1} - \bar{C}^{-1} k(X', X) \Sigma^R k(X, X') \bar{C}^{-1}\right] k(X', X) UU^\top \\
        &= \left[\bar{C}^{-1} - \bar{C}^{-1} k(X', X) \Sigma^R k(X, X') \bar{C}^{-1}\right] k(X', X) (RR^\top + NN^\top) \\
    \Sigma^R k(X, X') \bar{C}^{-1} k(X', X) R R^\top
        &= R \Sigma^{-1}R^\top  k(X, X') \bar{C}^{-1}  k(X', X) R R^\top \\
        &= R \Sigma^{-1} R^\top k(X, X') \bar{C}^{-1}  k(X', X) R R^\top\\
        &= R \Sigma^{-1} (\Sigma - (C_m^R)^{-1}) R^\top \\
        &= R(I - \Sigma^{-1} (C_m^R)^{-1}) R^\top
    \end{align*}
    So that
    \begin{align*}
        \tilde{C}^{-1} k(X', X) RR^\top &= \bar{C}^{-1} k(X', X) \left[ RR^\top - RR^\top + R\Sigma^{-1} (C_m^R)^{-1} R^\top \right] \\
        &= \bar{C}^{-1} k(X', X) R\Sigma^{-1} (C_m^R)^{-1} R^\top 
        \numberthis \label{eq:range_simpl}
    \end{align*}

    \paragraph{Mean Computation.} 
    Next we proceed to apply these results to compute the mean and covariance of \cref{eq:calib_req}.
    Note that Gaussianity is guaranteed by the fact that \cref{eq:calib_req} is an linear transformation of a difference of Gaussian random vectors, since the covariance is assumed to be independent of $y$.

    Considering the first term in \cref{eq:calib_split} we see that
    \begin{align*}
        \tilde{C}^{-1}(\bar{m} - f) &= (I - P) \bar{C}^{-1} (\bar{m} - f) \numberthis \label{eq:mean_simpl}\\
        \implies\mathbb{E}\left(\tilde{C}^{-1}(\bar{m} - f)\right)
        &= (I - P) \bar{C}^{-\frac{1}{2}} \mathbb{E} \left(\bar{C}^{-\frac{1}{2}}(\bar{m} - f) \right) \\
        &= 0
    \end{align*}
    due to the fact that the conditional GP is Bayesian and thus calibrated for the prior, by \cite[Example 1]{Cockayne2022Calib}.
    For the second term we have that
    \begin{align*}
        \tilde{C}^{-1}k(X', X)(\bar{v} - v) &= \tilde{C}^{-1} k(X', X) (RR^\top + NN^\top)(\bar{v} - v)
        \\
        &= \tilde{C}^{-1} k(X', X) RR^\top(\bar{v} - v)
    \end{align*}
    since, because the PLS is calibrated, $N^\top (\bar{v} - v) = 0$.
    Continuing, applying \cref{eq:range_simpl} we have
    \begin{align*}
        \tilde{C}^{-1} k(X', X) RR^\top(\bar{v} - v) &= \bar{C}^{-1} k(X', X) R \Sigma^{-1} (C_m^R)^{-1} R^\top (\bar{v} - v)\\
        \mathbb{E}\left(  \tilde{C}^{-1}k(X', X)(\bar{v} - v) \right) 
        &= \bar{C}^{-1} k(X', X) R \Sigma^{-1} (C_m^R)^{-\frac{1}{2}} \mathbb{E}\left((C_m^R)^{-\frac{1}{2}} R^\top (\bar{v} - v) \right)\\
        &= 0
    \end{align*}
    again due to calibratedness of the PLS.
    
    \paragraph{Variance Computation.}
    Next, for the variance, we have that
    \begin{align*}
    \mathbb{V}(\tilde{C}^{-\frac{1}{2}}(\tilde{m} - f)) &= \tilde{C}^{\frac{1}{2}} \bigg[
        \underbrace{\mathbb{V}\left(\tilde{C}^{-1} (\bar{m}-f)\right)}_{(1)}
        + \underbrace{\mathbb{V}\left(\tilde{C}^{-1} k(X', X) (\bar{v} - v) \right) }_{(2)} \\
    &\qquad\qquad - 2 \underbrace{\textup{Cov}\left(\tilde{C}^{-1}(\bar{m}-f), \tilde{C}^{-1} k(X', X)(\bar{v}-v)\right)}_{(3)} \bigg]\tilde{C}^{\frac{1}{2}} .
    \end{align*}
    Starting with $(1)$, from \cref{eq:mean_simpl} we obtain
    \begin{align*}
        \mathbb{V}(\tilde{C}^{-1}(\bar{m}-f)) &= (I-P) \bar{C}^{-\frac{1}{2}} \underbrace{\mathbb{V}\left[\bar{C}^{-\frac{1}{2}} (\bar{m} - f)\right]}_{=I} \bar{C}^{-\frac{1}{2}} (I-P)^\top \\
        &= (I-P) \bar{C}^{-1} (I-P)^\top
    \end{align*}
    where the fact that $\mathbb{V}\left[\bar{C}^{-\frac{1}{2}} (\bar{m} - f)\right] = I$ is due to calibratedness.
    Clearly
    \begin{align*}
        (I-P) \bar{C}^{-1} (I-P)^\top &= \bar{C}^{-1} - P\bar{C}^{-1} - \bar{C}^{-1} P^\top + P \bar{C}^{-1} P^\top \\
    \intertext{and}
        P \bar{C}^{-1} = \bar{C}^{-1} P^\top &= \bar{C}^{-1} k(X', X) \Sigma^{R} k(X, X') \bar{C}^{-1}.
    \end{align*}
    Furthermore,
    \begin{align*}
        P\bar{C}^{-1} P^\top &= \bar{C}^{-1} k(X', X) \Sigma^{R} k(X, X') \bar{C}^{-1} k(X', X) \Sigma^{R} k(X, X') \bar{C}^{-1} \\
        &= \bar{C}^{-1} k(X', X) R \Sigma^{-1}R^\top k(X, X') \bar{C}^{-1} k(X', X) R \Sigma^{-1} R^\top k(X, X') \bar{C}^{-1} \\
        &= \bar{C}^{-1} k(X', X) R \Sigma^{-1} \left[ \Sigma - (C_m^R)^{-1} \right] \Sigma^{-1} R^\top k(X, X') \bar{C}^{-1} \\
        &= \bar{C}^{-1} k(X', X) \Sigma^{R} k(X, X') \bar{C}^{-1} - \bar{C}^{-1} k(X', X) R\Sigma^{-1} (C_m^R)^{-1} \Sigma^{-1} R^\top k(X, X') \bar{C}^{-1}
    \end{align*}
    and so
    \begin{align*}
        \mathbb{V}(\tilde{C}^{-1}(m-f)) &= \bar{C}^{-1} - \bar{C}^{-1} k(X', X) \Sigma^{R} k(X, X') \bar{C}^{-1} - \bar{C}^{-1} k(X', X) R \Sigma^{-1} (C_m^R)^{-1} \Sigma^{-1} R^\top k(X, X') \bar{C}^{-1} \\
        &= \tilde{C}^{-1} - \bar{C}^{-1} k(X', X) R \Sigma^{-1} (C_m^R)^{-1} \Sigma^{-1} R^\top k(X, X') \bar{C}^{-1}.
        \numberthis \label{eq:term_1_simpl}
    \end{align*}
    Now for $(2)$, again applying \cref{eq:range_simpl}
    \begin{align*}
        \mathbb{V}\left(\tilde{C}^{-1}k(X', X)(\bar{v} - v)\right) &= \mathbb{V}\left(\tilde{C}^{-1} k(X', X) (RR^\top + NN^\top)(\bar{v} - v) \right) \\
        &= \mathbb{V}\left(\tilde{C}^{-1} k(X', X) RR^\top (\bar{v} - v)\right)
    \end{align*}
    since $N^\top (\bar{v} - v) = 0$ due to calibratedness of the PLS.
    Further we have
    \begin{align*}
        \mathbb{V}\left(\tilde{C}^{-1} k(X', X) RR^\top (\bar{v} - v)\right) &= \mathbb{V}\left(\bar{C}^{-1} k(X', X) R \Sigma^{-1} (C_m^R)^{-1} R^\top (\bar{v}-v) \right) \\
        &= \bar{C}^{-1} k(X', X) R \Sigma^{-1} (C_m^R)^{-\frac{1}{2}} \mathbb{V} \left((C_m^R)^{-\frac{1}{2}} R^\top (\bar{v}-v)\right) (C_m^R)^{-\frac{1}{2}} \Sigma^{-1} R^\top k(X, X') \bar{C}^{-1} \\
        &= \bar{C}^{-1} k(X', X) R \Sigma^{-1} (C_m^R)^{-1} \Sigma^{-1} R^\top k(X, X') \bar{C}^{-1}
    \end{align*}
    where the inner variance on the second line is $I$ again due to calibratedness.
    This cancels with \cref{eq:term_1_simpl}, yielding
    \begin{align*}
        \mathbb{V}(\tilde{C}^{-1}(m-f)) + \mathbb{V}(\tilde{C}^{-1}k(X', X)(\bar{v} - v)) &= \bar{C}^{-1} - \bar{C}^{-1} k(X', X) \Sigma^{R} k(X, X') \bar{C}^{-1} \\
        &= \tilde{C}^{-1}
    \end{align*}
    so that
    \begin{align}
        \tilde{C}^{\frac{1}{2}} \left[\mathbb{V}(\tilde{C}^{-1}(m-f)) + \mathbb{V}(\tilde{C}^{-1}k(X', X)(\bar{v} - v)) \right] \tilde{C}^{\frac{1}{2}}
        &= I \label{eq:already_have_identity}
    \end{align}

    Finally, we examine the cross covariance term $(3)$.
    Clearly
    \begin{align*}
        \textup{Cov}\left(\tilde{C}^{-1}(\bar{m}-f), \tilde{C}^{-1} k(X', X)(\bar{v}-v)\right) &= \tilde{C}^{-1} \textup{Cov}\left(\bar{m} - f, \bar{v}-v\right) k(X, X') \tilde{C}^{-1}.
    \end{align*}
    and due to bilinearity of the covariance we have
    \begin{align*}
        \textup{Cov}\left(\bar{m} - f, \bar{v}-v\right) &= \textup{Cov}\left(\bar{m} - f, \bar{v}\right) - \textup{Cov}\left(\bar{m}, v\right) + \textup{Cov}\left(f, v\right) .
    \end{align*}
    The last two terms can be calculated directly.
    Since $\bar{m}(X') = m_0(X') + k(X', X)v$ and $v = \gramian^{-1} y$, we get
    \begin{align*}
        \textup{Cov}\left(\bar{m}, v\right) &= k(X', X)\textup{Cov}\left(v, v\right) \\
        &= k(X', X) \gramian^{-1} \\
        \textup{Cov}\left(f, v\right) &= \textup{Cov}(f,y) \gramian^{-1} \\
        &= k(X', X) \gramian^{-1} 
    \end{align*}
    and so
    \begin{align*}
        \textup{Cov}\left(\bar{m} - f, \bar{v}-v\right) &= \textup{Cov}\left(\bar{m} - f, \bar{v}\right) - \gramian^{-1} k(X', X) + \gramian^{-1} k(X', X)  \\
        &= \textup{Cov}\left(\bar{m} - f, \bar{v}\right)
    \end{align*}
    We can also simplify this result again using the expressions for $\bar{m}$ and $f$.
    Since $\bar{m}(X') = m_0(X') + K(X', X) \gramian^{-1} y$ we have 
    \begin{align*}
        \textup{Cov}\left(\bar{m} - f, \bar{v}\right) &= k(X', X) \gramian^{-1}\textup{Cov}(y, \bar{v}) - \textup{Cov}(f(X'), \bar{v})\\
        &= 0
    \end{align*}
    by condition \cref{cond:cross} from the theorem.

    Putting this together we obtain that
    \begin{align*}
        \tilde{C}^{\frac{1}{2}} \left[\mathbb{V}(\tilde{C}^{-1}(\tilde{m}-f))\right] \tilde{C}^{\frac{1}{2}}
        &= I
    \end{align*}
    as required.
\end{proof}

\begin{proof}[Proof of \cref{corr:cross}]
    Since $\bar{v} = My + g$, 
    \begin{align*}
        \textup{Cov}(f(X'), \bar{v}) &= \textup{Cov}(f(X'), y) M^\top \\
        &= k(X', X) M^\top 
    \end{align*}
    and 
    \begin{align*}
        \textup{Cov}(y, v) &= \textup{Cov}(y, My) \\
        &= \gramian M^\top
    \end{align*}
    so that
    \begin{align*}
        K(X', X) \gramian^{-1} \textup{Cov}(y, \bar{v}) &= K(X', X) M^\top ,
    \end{align*}
    completing the proof.
\end{proof}

\begin{proof}[Proof of \cref{prop:pim_itergp}]
    We first have that $C_0 = \gramian^{-1}$. Therefore,
    \begin{align*}
        C_1 &= \itermat \gramian^{-1} \itermat^\top \\
        &= (I - \ccitermat\gramian) \gramian^{-1} (I - \ccitermat\gramian)^\top \\
        &= \gramian^{-1} - \ccitermat - \ccitermat^\top + \ccitermat\gramian\ccitermat^\top
    \end{align*}
    so that $C_1 = \gramian^{-1} - D_1$ where $D_1 = \ccitermat + \ccitermat^\top - \ccitermat\gramian\ccitermat^\top$. 
    
    Proceeding inductively, suppose that $C_{i-1} = \gramian^{-1} - D_{i-1}$. Then applying the above,
    \begin{align*}
        C_i &= (I - \ccitermat\gramian) (\gramian^{-1} - D_{i-1}) (I - \ccitermat\gramian)^\top \\
        &= \gramian^{-1} - \ccitermat - \ccitermat^\top + \ccitermat\gramian\ccitermat^\top \\
        &\qquad- (I - \ccitermat\gramian) D_{i-1} (I - \ccitermat\gramian)^\top \\
        &= \gramian^{-1} - D_{i}
    \end{align*}
    where 
    \begin{equation}
        D_{i} = \ccitermat + \ccitermat^\top - \ccitermat\gramian\ccitermat^\top + (I - \ccitermat\gramian) D_{i-1} (I - \ccitermat\gramian)^\top.
    \end{equation}
    We therefore obtain the required structure. 
\end{proof}

\begin{proof}[Proof of \cref{prop:gs_properties}]
    Note that since $C_m$ is a Gramian matrix, $\textup{null}(C_m) = \textup{null}(G^{-\frac{1}{2}} (M^m)^\top) = \textup{null}((M^m)^\top)$, since $G^{-\frac{1}{2}}$ is positive definite.
    (This follows from the fact that if $C_m v = 0$ then $\norm{C_m^\frac{1}{2} v}_2 = 0$ for any factor $C_m^\frac{1}{2}$).
    
    For Gauss-Seidel with $m=1$ we have that $\textup{rank}(M^\top) = \textup{rank}(U^\top L^{-1})$.
    From \cite[Fact 6.3]{Ipsen}, the range of $U^\top L^{-\top}$ is the same as the range of $U^\top$, which is easily seen to be $\textup{span}(e_2, \dots, e_d)$ thanks to the strict lower triangular structure of $U^\top$.
    As a result the rank of $M^\top$ is $d-1$.
    Using the rank-nullity theorem, we therefore have that the null space of $M^\top$ is $1$-dimensional, and it is similarly easy to see that the null space must be $\textup{span}(L^\top e_d)$.

    Proceeding to $m=2$, we will identify the null space of $(M^\top)^2$.
    Consider $(M^\top)^2 v$ for arbitrary $v$.
    Clearly $(M^\top)^2 v = 0$ if for some $\alpha$ either:
    \begin{enumerate}
        \item $v = \alpha L^\top e_d$ (i.e.\ $v$ lies in the null space of $M^\top$).
        \item $M^\top v = \alpha L^\top e_d$ (i.e.\ $M^\top v$ lies in the null space of $M^\top$).
    \end{enumerate}
    Considering the latter, if $M^\top v = L^\top e_d$ then $Mv$ is equal to the last column of $L^\top$.
    However since the last column of $L^\top$ is dense, it does not lie in the range of $U^\top L^{-\top}$ (since the first component is nonzero).
    Hence, the null space of $(M^\top)^2$ is the same as the null space of $M^\top$, and its rank is $d-1$ using the rank-nullity theorem.
    Iterating this argument shows that the null space of $(M^m)^\top$ is $\textup{span}(L^{\top} e_d)$.
    The statement about ranks again follows from the rank-nullity theorem.
    This completes the proof.
\end{proof}

\section{Simulation-Based Calibration} \label{sec:sbc}

In this section we outline the simulation-based calibration procedure introduced in \citet{Talts2018}, which can be used to test for calibratedness numerically.
The approach operates on similar principles to those described in \cref{sec:calib}, but pushes samples through a test functional to produce samples whose distribution can be more easily evaluated empirically.
As a result, these tests are a necessary condition for strong calibration but not a sufficient one.

Since we operate in a Gaussian framework we will use a test statistic derived from projecting the distribution through a vector $w^\top$, as the required marginal distribution is then straightforward to derive.
In this setting the simulation-based calibration test reduces to that described in \cref{alg:sbc}.

\begin{algorithm}
    \caption{Simulation-Based Calibration. $\Phi$ denotes the CDF of the standard Gaussian distribution.} \label{alg:sbc}
    \begin{algorithmic}[1]
    \Require Prior $\mu_0 = \mathcal{N}(u_0, \Sigma_0)$, data-generating model $\textup{\textsc{dgm}}$, learning procedure $\mu$, number of simulations $N_\textsc{sim}$, test vector $w$.
    \For{$i = 1$ to $N_\textsc{sim}$}
        \State $u_i^\star \sim \mu_0$
        \State $y_i = \textup{\textsc{dgm}}(u_i^\star)$
        \State $\mu_i = \mu(\mu_0, y_i) = \mathcal{N}(\bar{u}_i, \bar{\Sigma}_i)$
        \State $t_i = \Phi\left(\frac{w^\top (\bar{u}_i - u_i^\star)}{(w^\top \Sigma_i w)^\frac{1}{2}}\right)$
    \EndFor
    \State Plot histogram of $\{t_i\}_{i=1}^N$ and compare to $\mathcal{U}(0,1)$.
    \end{algorithmic}
\end{algorithm}

Typically, we will choose $w$ to be a random unit vector.
\cref{alg:sbc} can then be used to check whether an arbitrary Gaussian learning procedure is calibrated and, in particular, to highlight miscalibration of CAGP-CG in \cref{sec:simulations}.
In the event of miscalibration, the histogram may have a U-shape in the event of an overconfident posterior (i.e.\ the truth is typically in the tails of the learned distribution) or an inverted U-shape for a conservative posterior (i.e.\ the truth is typically near the modal point), though of course other shapes are possible.

\section{Further Simulation Results}
\subsection{Synthetic Problem} \label{sec:sup-synthetic}

\begin{figure}
    \begin{subfigure}{0.5\textwidth}
    
    \includegraphics[width=\textwidth]{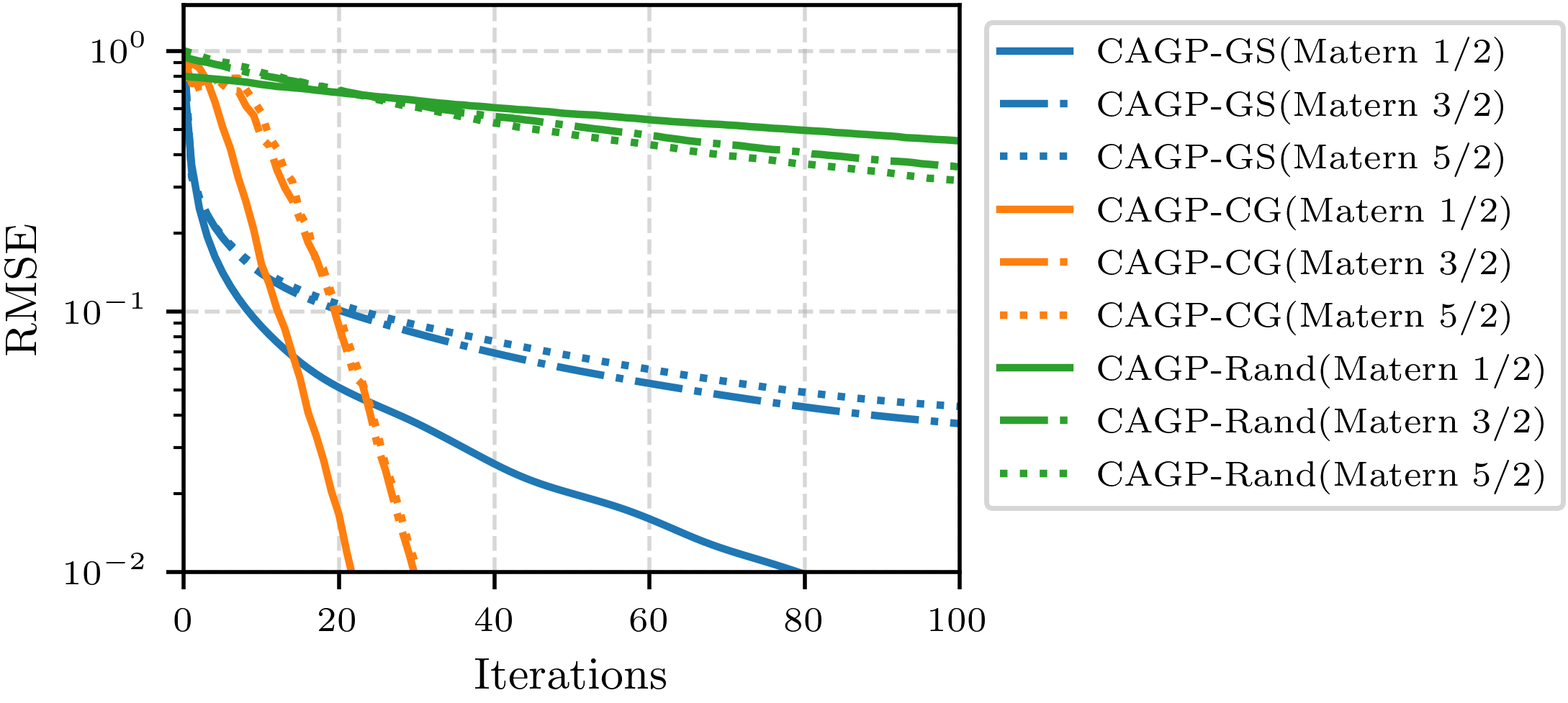}
    \caption{RMSE} \label{fig:synthetic:rmse_vary_smoothness}
    \end{subfigure}
    \begin{subfigure}{0.5\textwidth}
    \includegraphics[width=\textwidth]{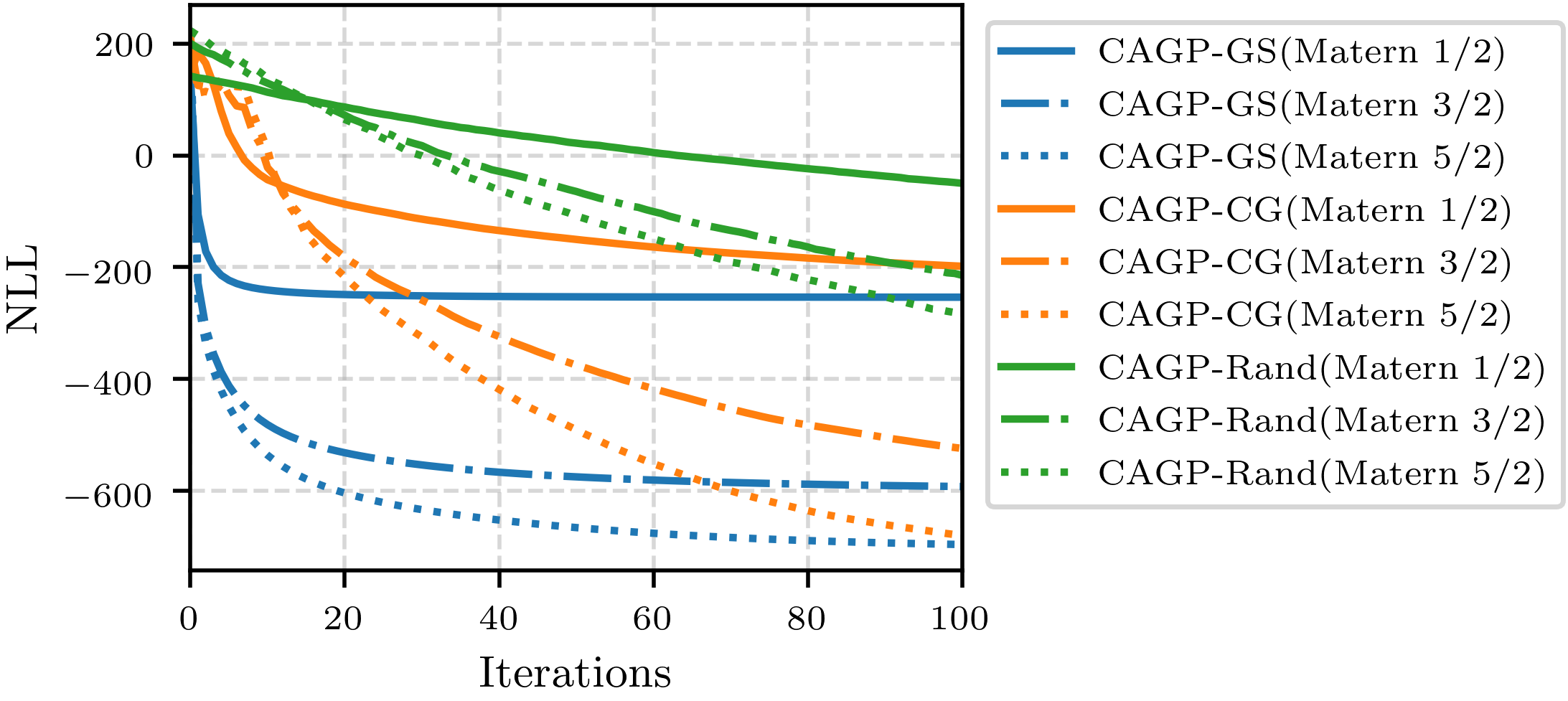}
    \caption{NLL} \label{fig:synthetic:nll_vary_smoothness}
    \end{subfigure}
    \caption{Convergence behaviour for the synthetic problem in \cref{sec:synthetic}.} \label{fig:synthetic:vary_smoothness}
    \end{figure}

\cref{fig:synthetic:vary_smoothness} shows that the promising behaviour of \ac{CAGP}-GS \ac{RMSE} and \ac{NLL} extend to rougher and smoother processes. Here we plot \ac{RMSE} and \ac{NLL} for Matèrn $\sfrac{3}{2}$, $\sfrac{3}{2}$ and $\sfrac{5}{2}$ for a fixed length scale of $l = 0.2$ for the same setup in \cref{sec:synthetic}. \ac{CAGP}-GS outperforms \ac{CAGP}-CG in \ac{RMSE} and \ac{NLL} for initial few iterations as expected.

\begin{figure*}
    \begin{subfigure}{0.32\textwidth}
        \includegraphics[width=\textwidth]{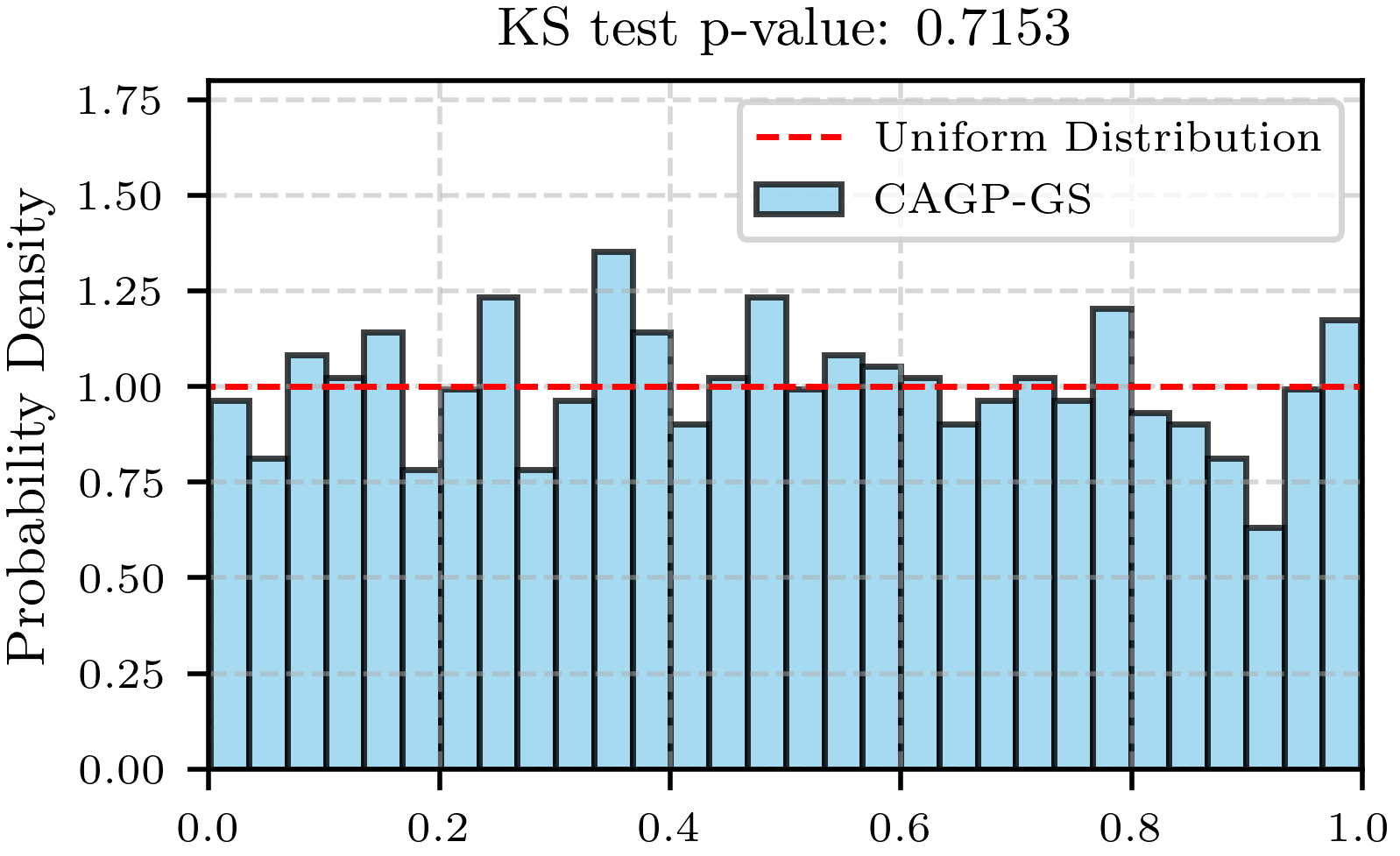}
        \caption{CAGP-GS at $m = 50$} \label{fig:synthetic:sbc-gs50}
    \end{subfigure}
    \begin{subfigure}{0.32\textwidth}
        \includegraphics[width=\textwidth]{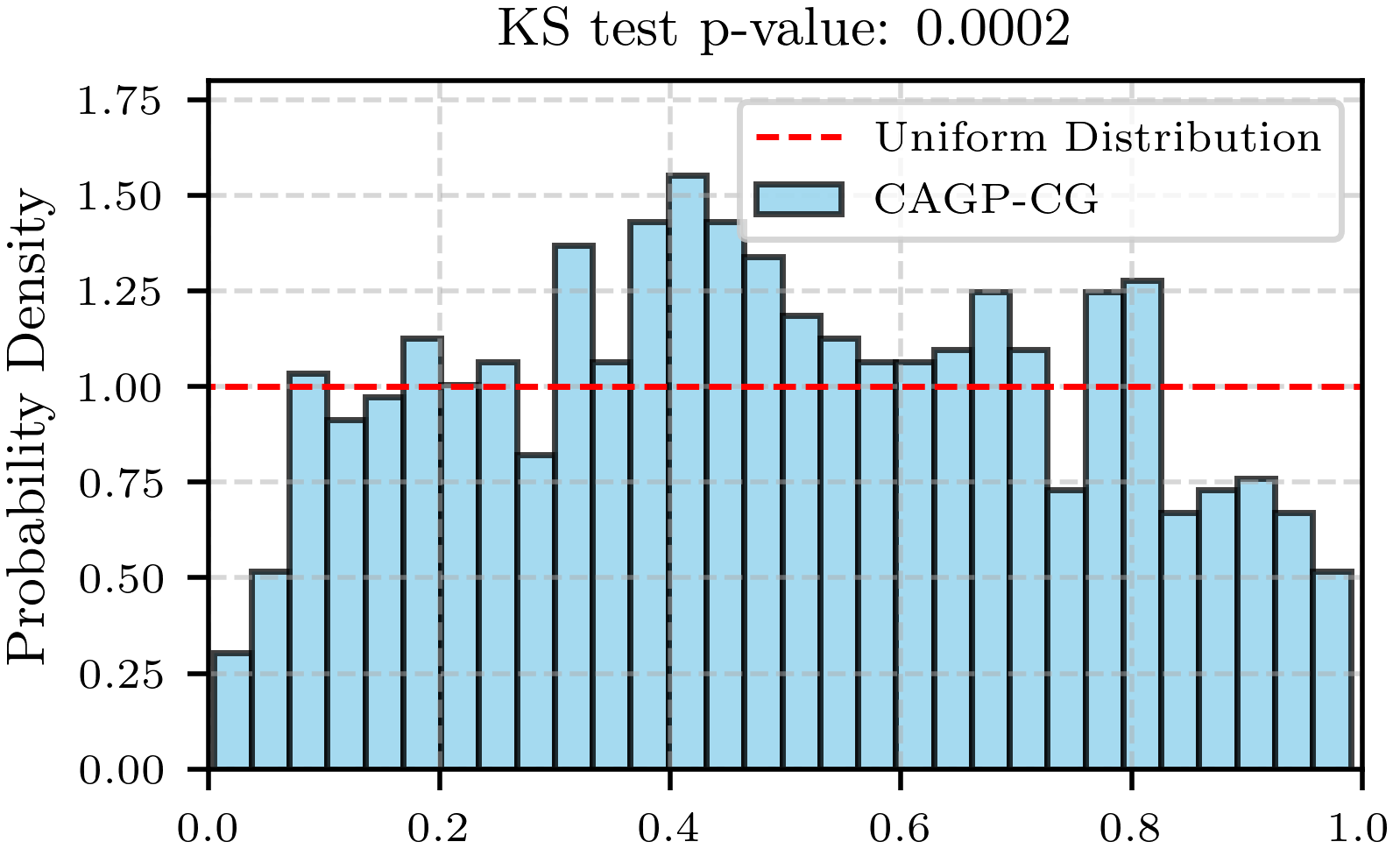}
        \caption{CAGP-CG at $m = 50$} \label{fig:synthetic:sbc-cg50}
    \end{subfigure}
    \begin{subfigure}{0.32\textwidth}
        \includegraphics[width=\textwidth]{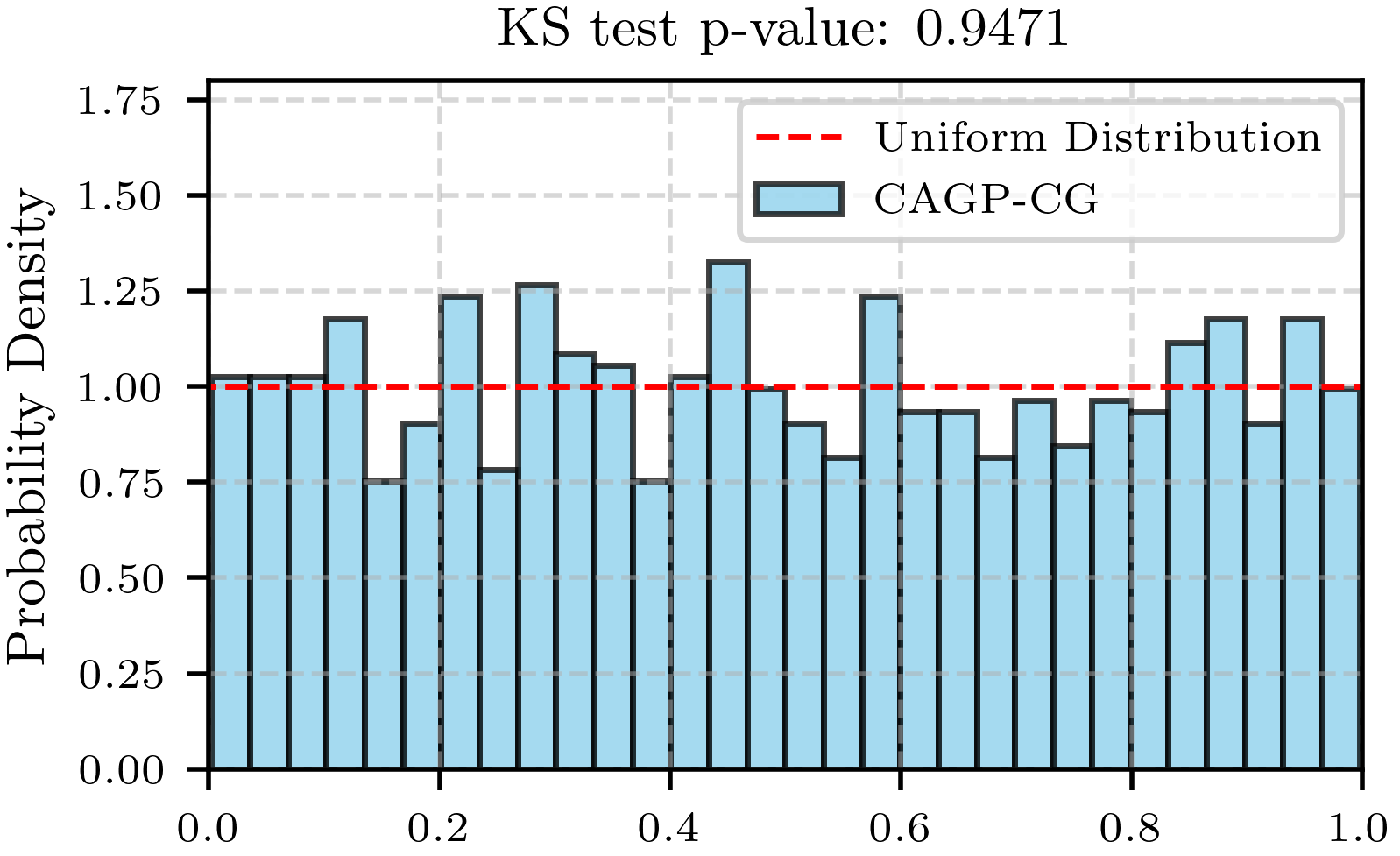}
        \caption{CAGP-CG at $m = 100$} \label{fig:synthetic:sbc-cg100}
    \end{subfigure}
    \caption{Simulation-based calibration results for the synthetic problem described in \cref{sec:synthetic} for varying number of iterations} \label{fig:synthetic:sbc_m}
    \end{figure*}

In \cref{fig:synthetic:sbc_m} we show the effect of increasing number of iterations on calibration of \ac{CAGP}-GS and \ac{CAGP}-CG. We can see that while \ac{CAGP}-GS remains calibrated (\cref{fig:synthetic:sbc-gs50}), \ac{CAGP}-CG  becomes well-calibrated with increasing $m$. This is because in \ac{CAGP}-CG we get linear convergence of the covariance but superlinear convergence of the mean, leading to poor calibration. Nevertheless, eventually the covariance ``catches up'' to the mean, and the posterior becomes calibrated for high iterations (\cref{fig:synthetic:sbc-cg100}).

\begin{figure*}
    \begin{subfigure}{0.5\textwidth}
        \includegraphics[width=\textwidth]{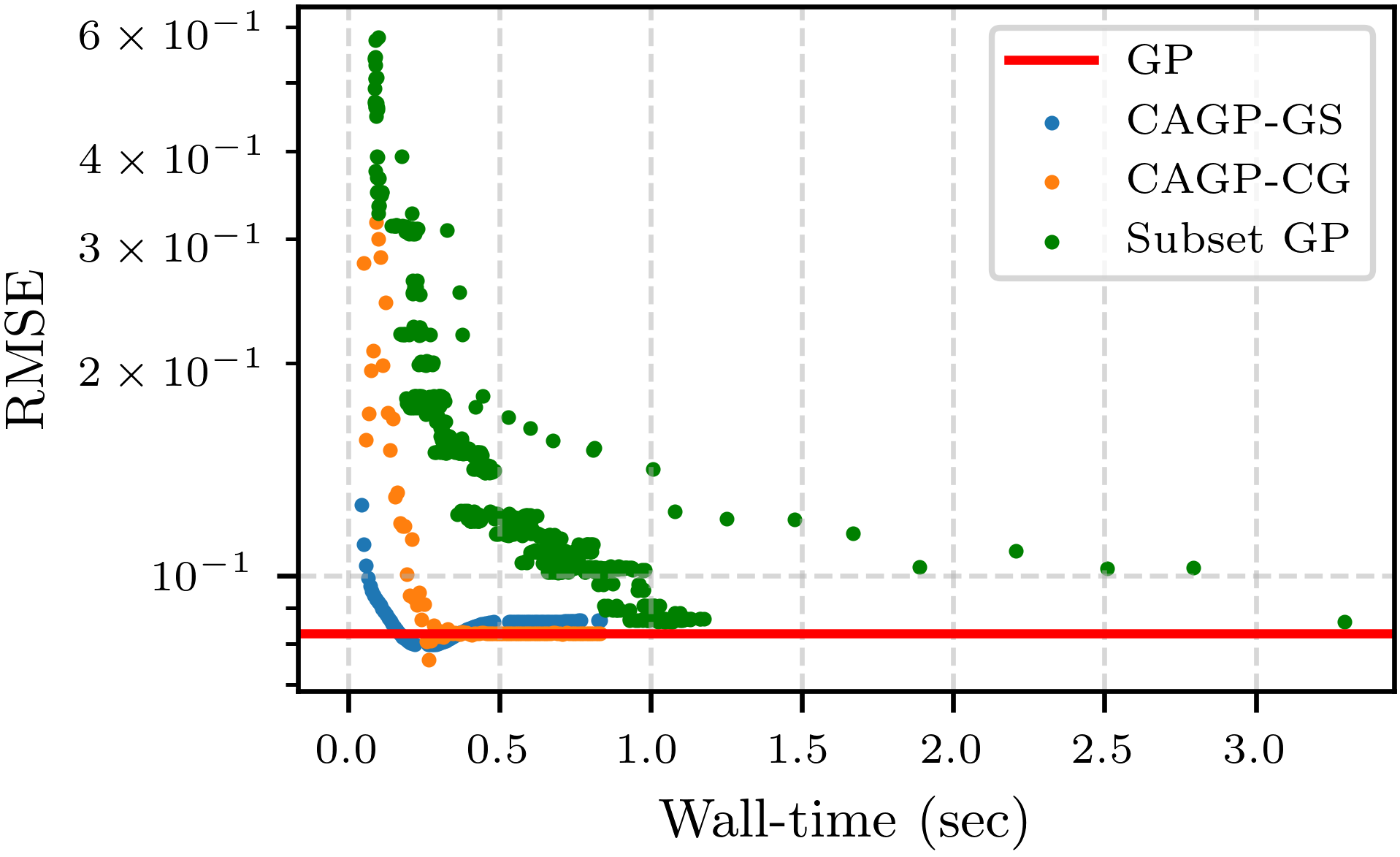}
        \caption{RMSE against time for \cref{sec:synthetic}}
        \label{fig:synthetic:time_rmse}
    \end{subfigure}
    \begin{subfigure}{0.5\textwidth}
        \includegraphics[width=\textwidth]{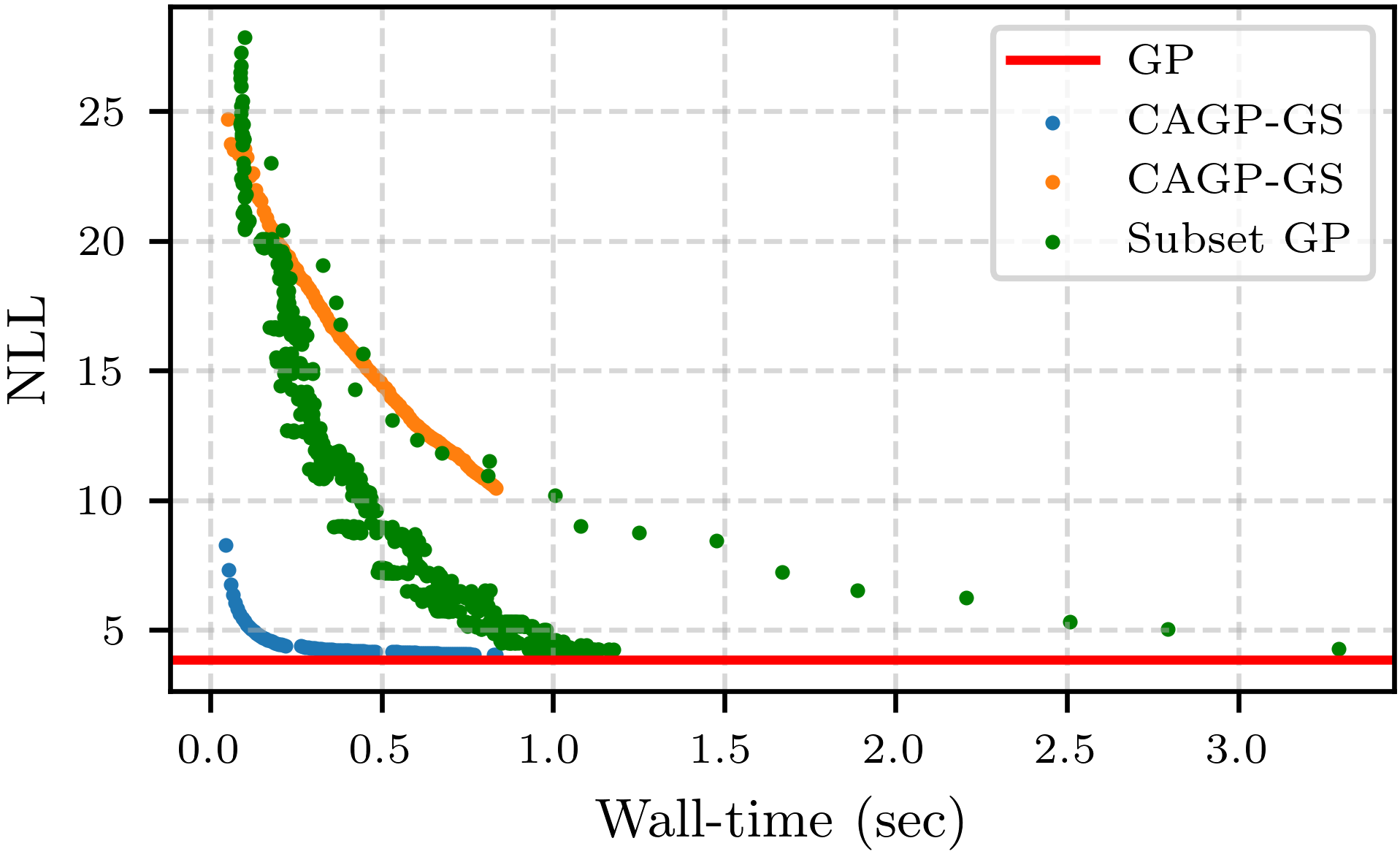}
        \caption{NLL against time for \cref{sec:synthetic}}
        \label{fig:synthetic:time_nll}
    \end{subfigure}
    \caption{RMSE and NLL against time for \cref{sec:synthetic}}
    \label{fig:synthetic:time}
\end{figure*}

Another approach to reducing the cost of \ac{GP} approximation is to use a subset of the training set to fit the regression model. 
In \cref{fig:synthetic:time} we compare the performance of \ac{CAGP}-GS and \ac{CAGP}-CG with this subset GP approach, in terms of time taken to compute RMSE and NLL by increasing the number of iterations for \ac{CAGP}-CG and \ac{CAGP}-CG, and increasing size of a randomly chosen subset for subset GP. 
We use a setting where \ac{CAGP}-GS is presumed to work well, with $d = 100$ and $d_\textsc{test} = 25$. We use the \texttt{numpy} implementation of Cholesky factorisation to compute the GP mean and posterior for the subset GP, and \texttt{jax} to jit compile our implementations of all three methods for speed.
From \cref{fig:synthetic:time_rmse}, we can see that \ac{CAGP}-GS and \ac{CAGP}-CG converge faster than subset GP, with \ac{CAGP}-GS being the fastest for the first few iterations. With its better calibrated posterior, \ac{CAGP}-GS also outperforms the other two approaches in NLL. However, it should be noted that this performance heavily depends on implementation, as well as the hardware used.
For example, the \texttt{jax} Cholesky implementation is a parallelisable block algorithm, whereas our Gauss-Seidel implementation is harder to parallelise as noted in \cref{sec:conclusion}.
Nevertheless the results appeared to be similar across a range of hardware; the figures presented were from a machine equipped with an i5-1245U CPU with 10 cores, but similar performance was observed when using the University of Southampton's Iridris5 cluster on a single node with 40 tasks.
    
\subsection{ERA5 Regression Problem}

\begin{figure*}
    \begin{subfigure}{0.32\textwidth}
        \includegraphics[width=\textwidth]{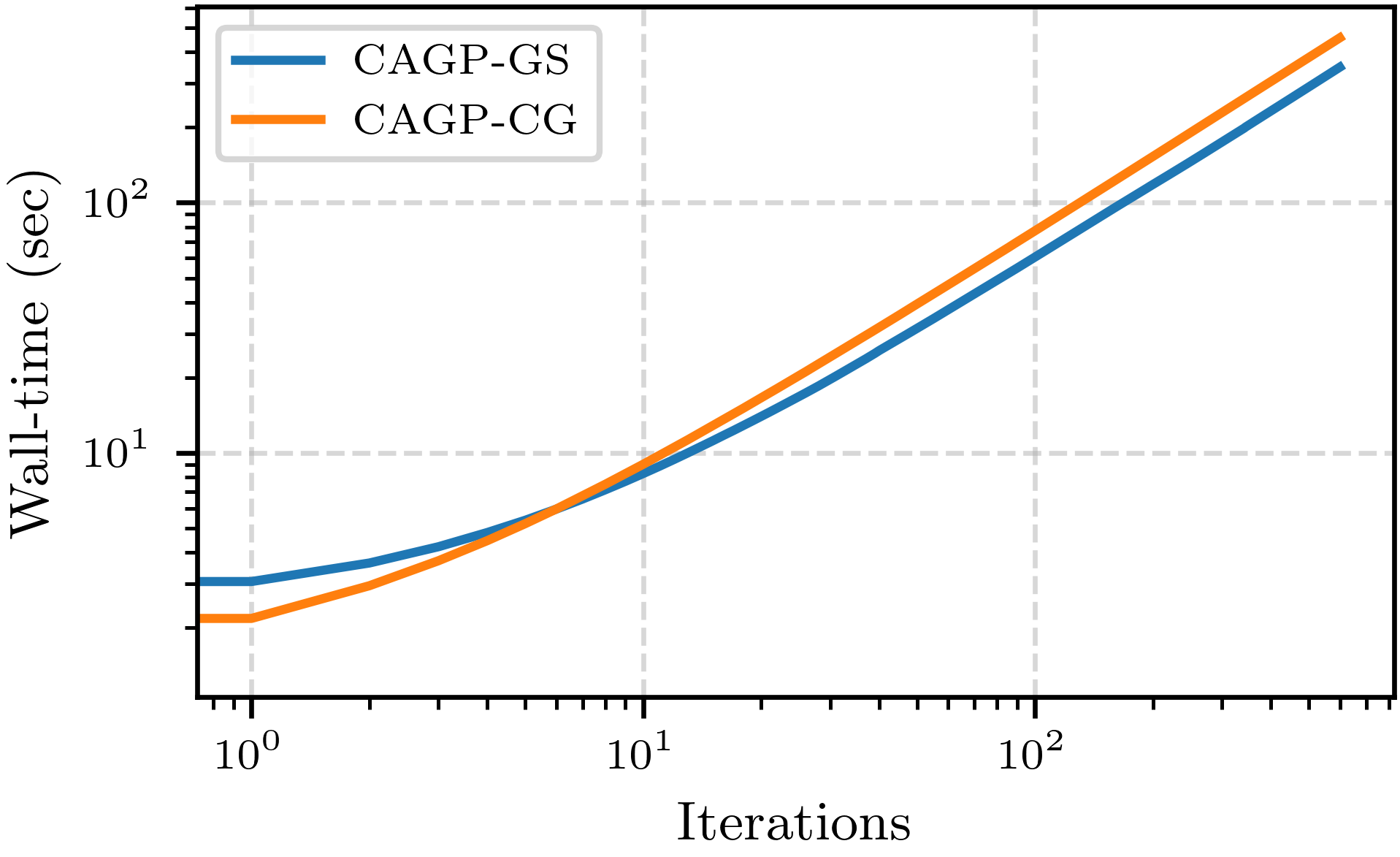}
        \caption{Varying $m$.}
    \end{subfigure}
    \begin{subfigure}{0.32\textwidth}
        \includegraphics[width=\textwidth]{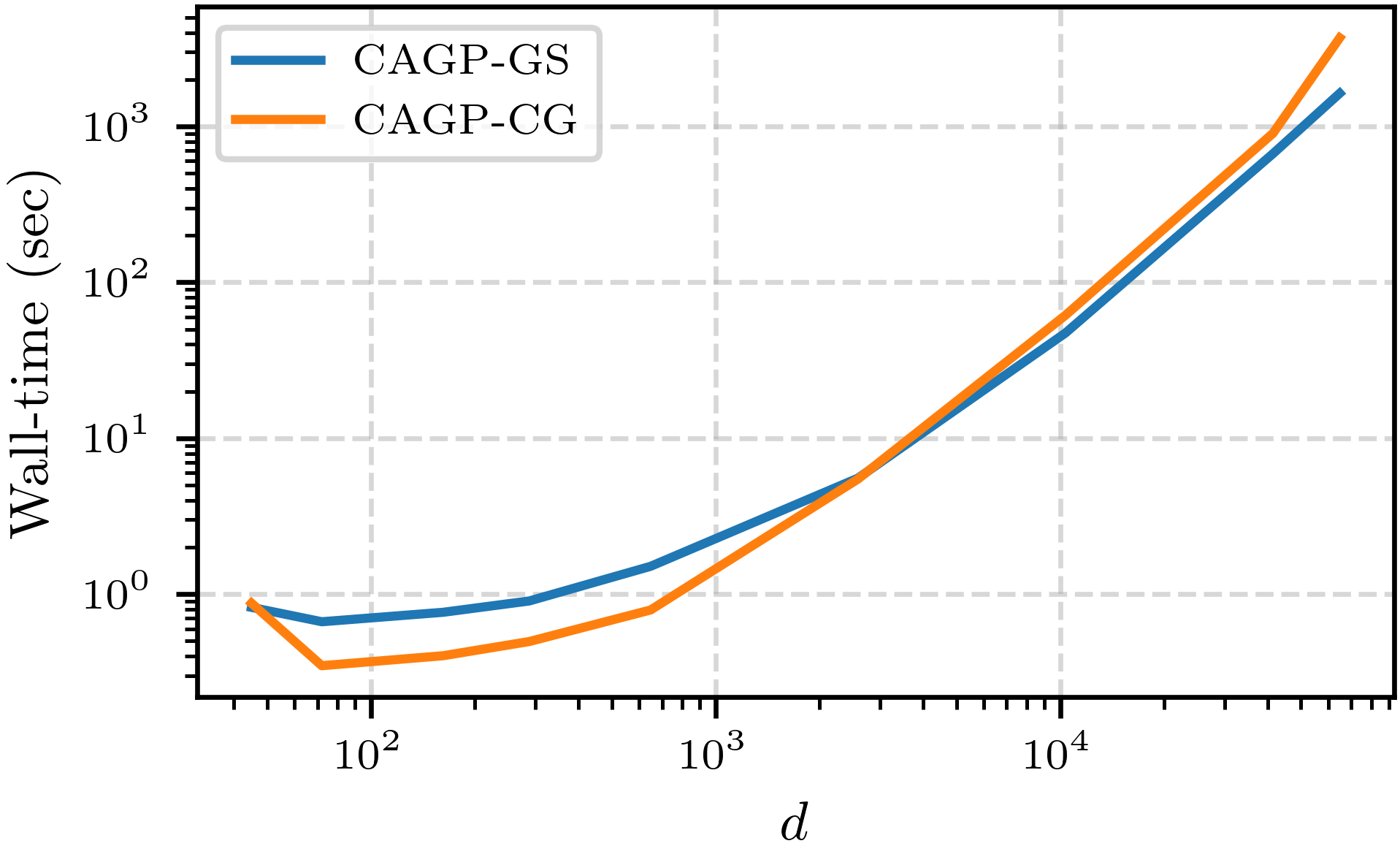}
        \caption{Varying $d$.}
    \end{subfigure}
    \begin{subfigure}{0.32\textwidth}
        \includegraphics[width=\textwidth]{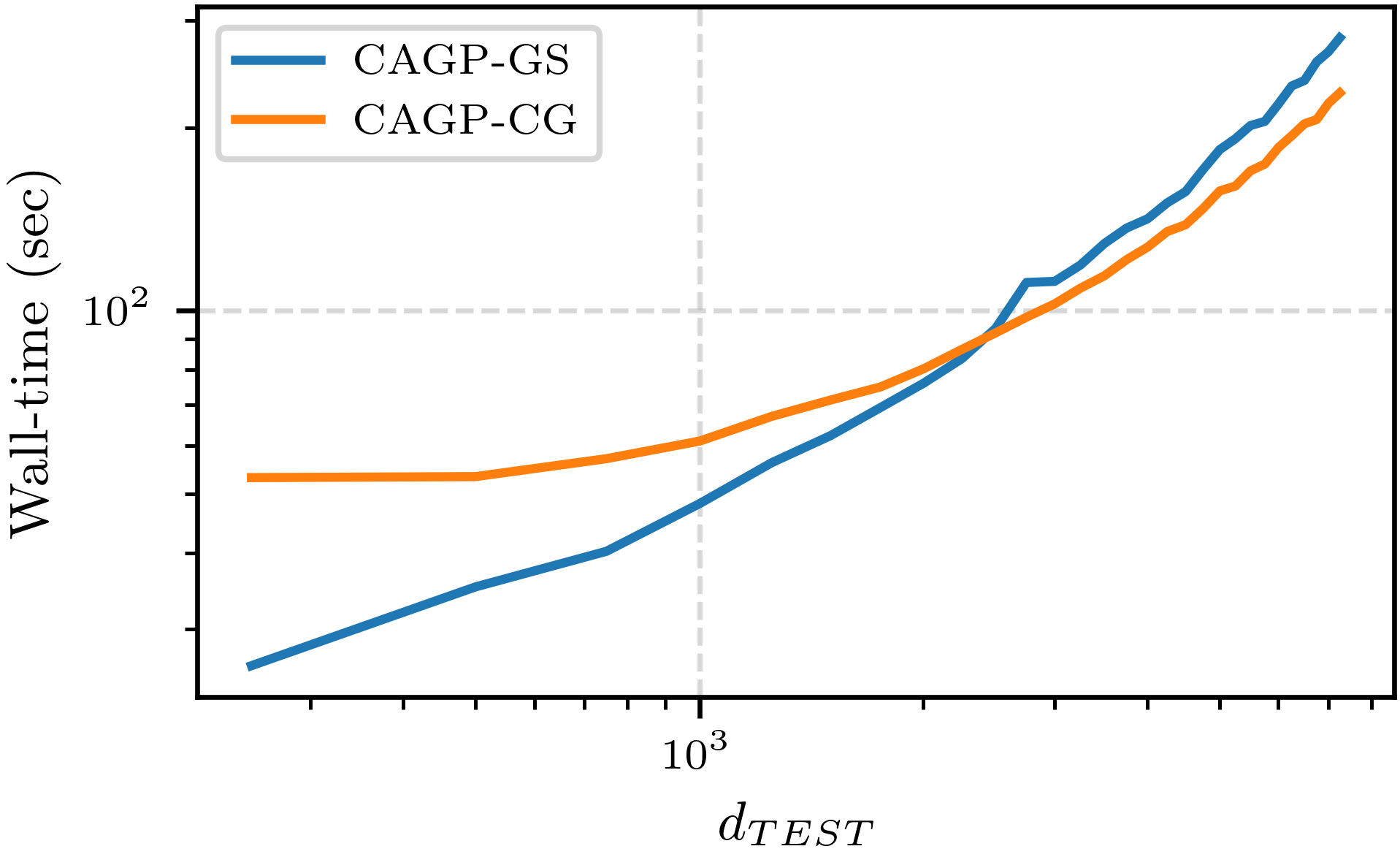}
        \caption{Varying $d_\textsc{test}$.}
    \end{subfigure}
    \caption{Timings for the large-scale regression problem from \cref{sec:era5} as a variety of parameters are varied.} \label{fig:era5:timings}
\end{figure*}

\begin{figure*}
    \begin{subfigure}{0.5\textwidth}
        \includegraphics[width=\textwidth]{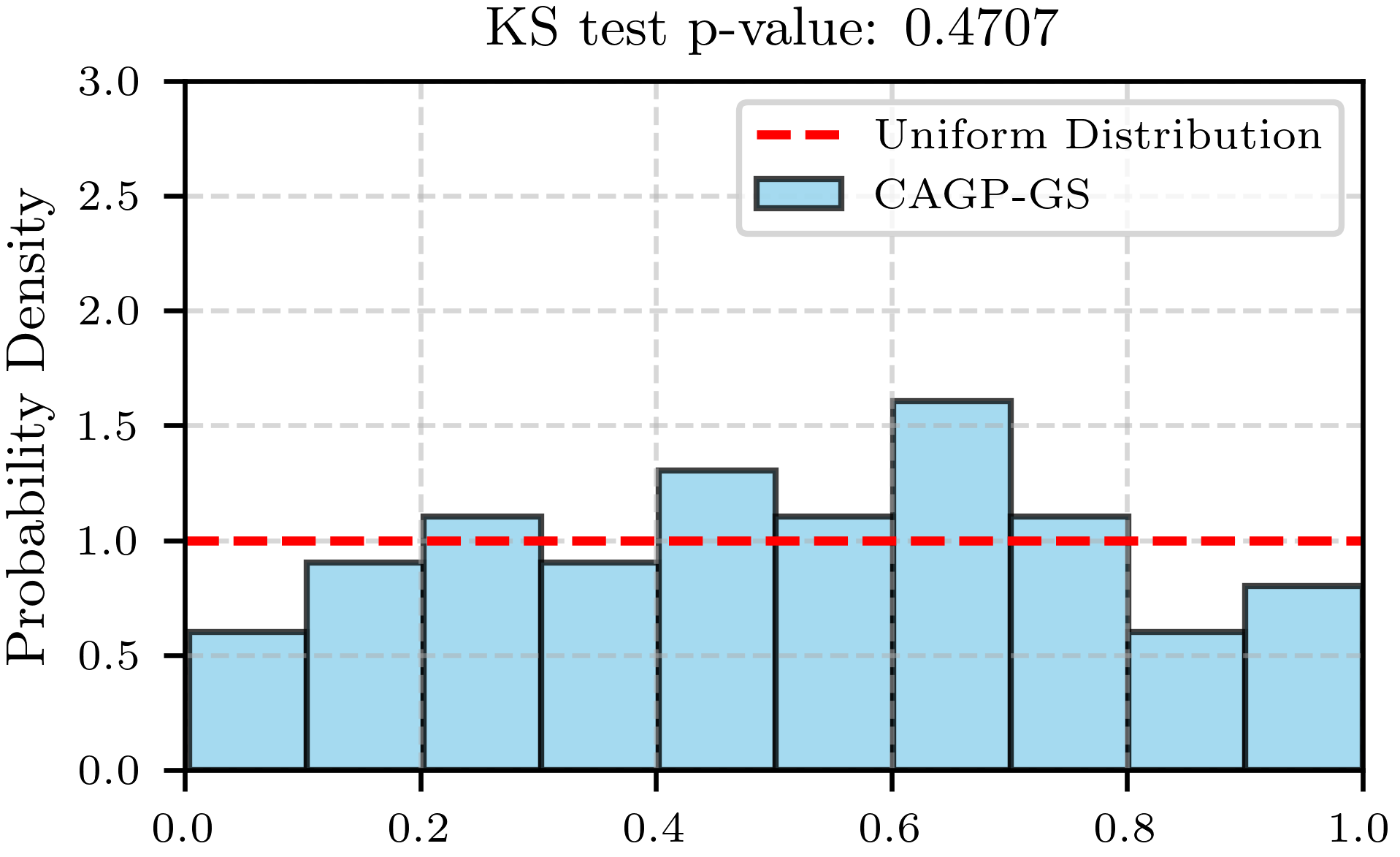}
        \caption{CAGP-GS}
    \end{subfigure}
    \begin{subfigure}{0.5\textwidth}
        \includegraphics[width=\textwidth]{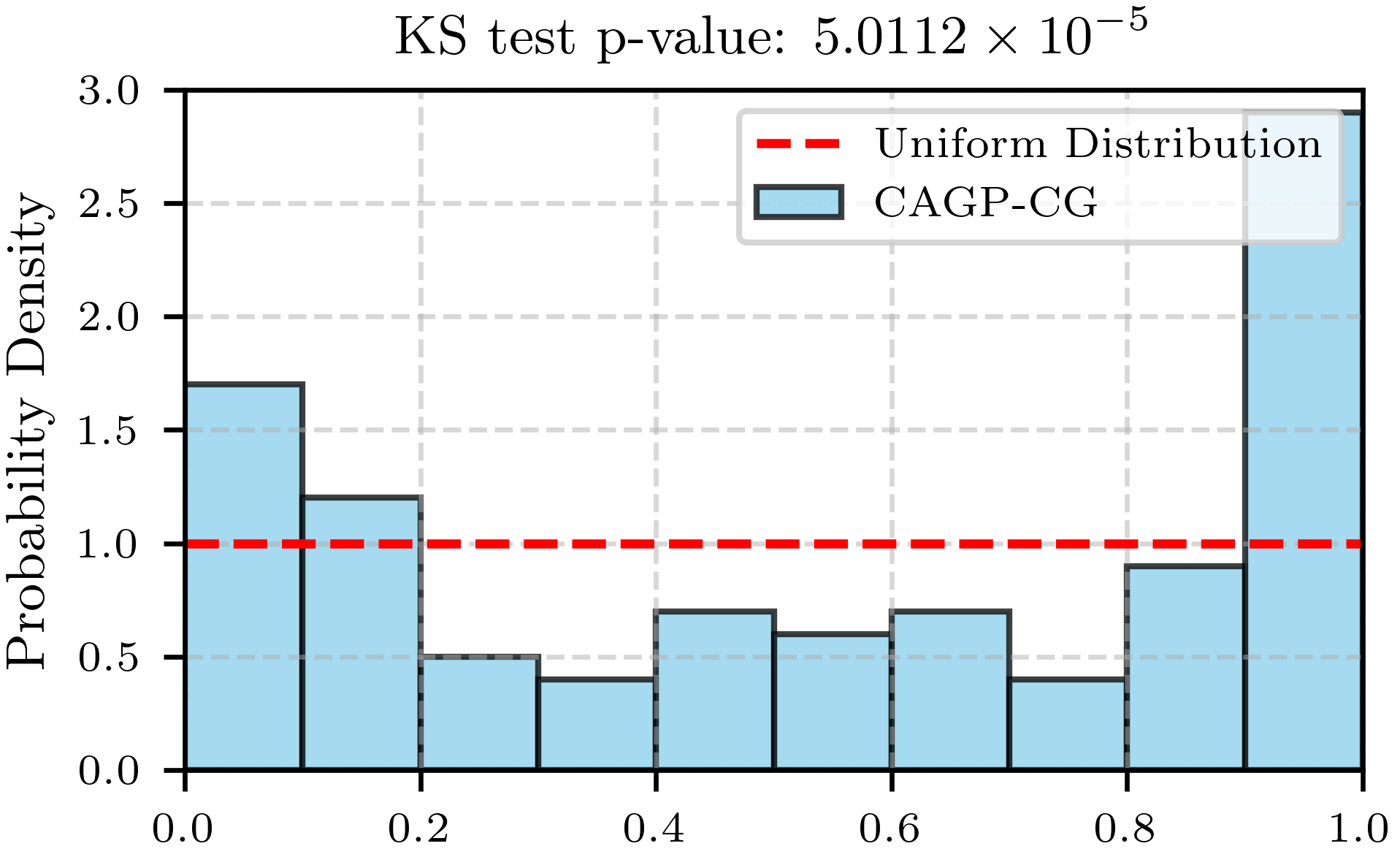}
        \caption{CAGP-CG}
    \end{subfigure}
    \caption{Calibration results for the large-scale regression problem from \cref{sec:era5}} \label{fig:era5:zscores}
\end{figure*}

\cref{fig:era5:timings} reports timings for the ERA5 regression problem from \cref{sec:era5}, while \cref{fig:era5:zscores} shows results of the calibratedness experiment reported therein.

\begin{figure*}
    \begin{subfigure}{0.5\textwidth}
        \includegraphics[width=\textwidth]{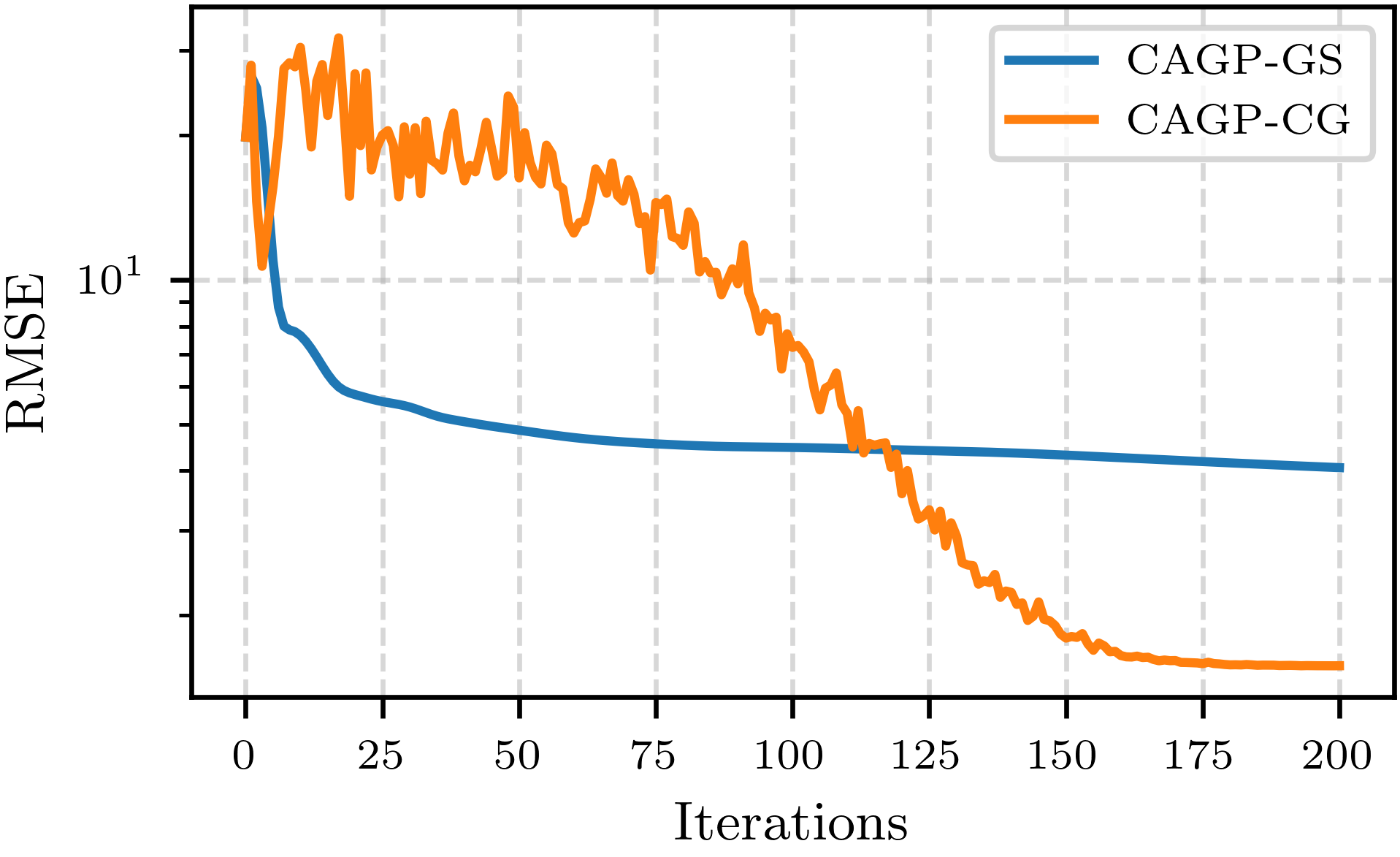}
        \caption{RMSE for the large-scale regression from \cref{sec:era5}}
        \label{fig:era5:rmse}
    \end{subfigure}
    \begin{subfigure}{0.5\textwidth}
        \includegraphics[width=\textwidth]{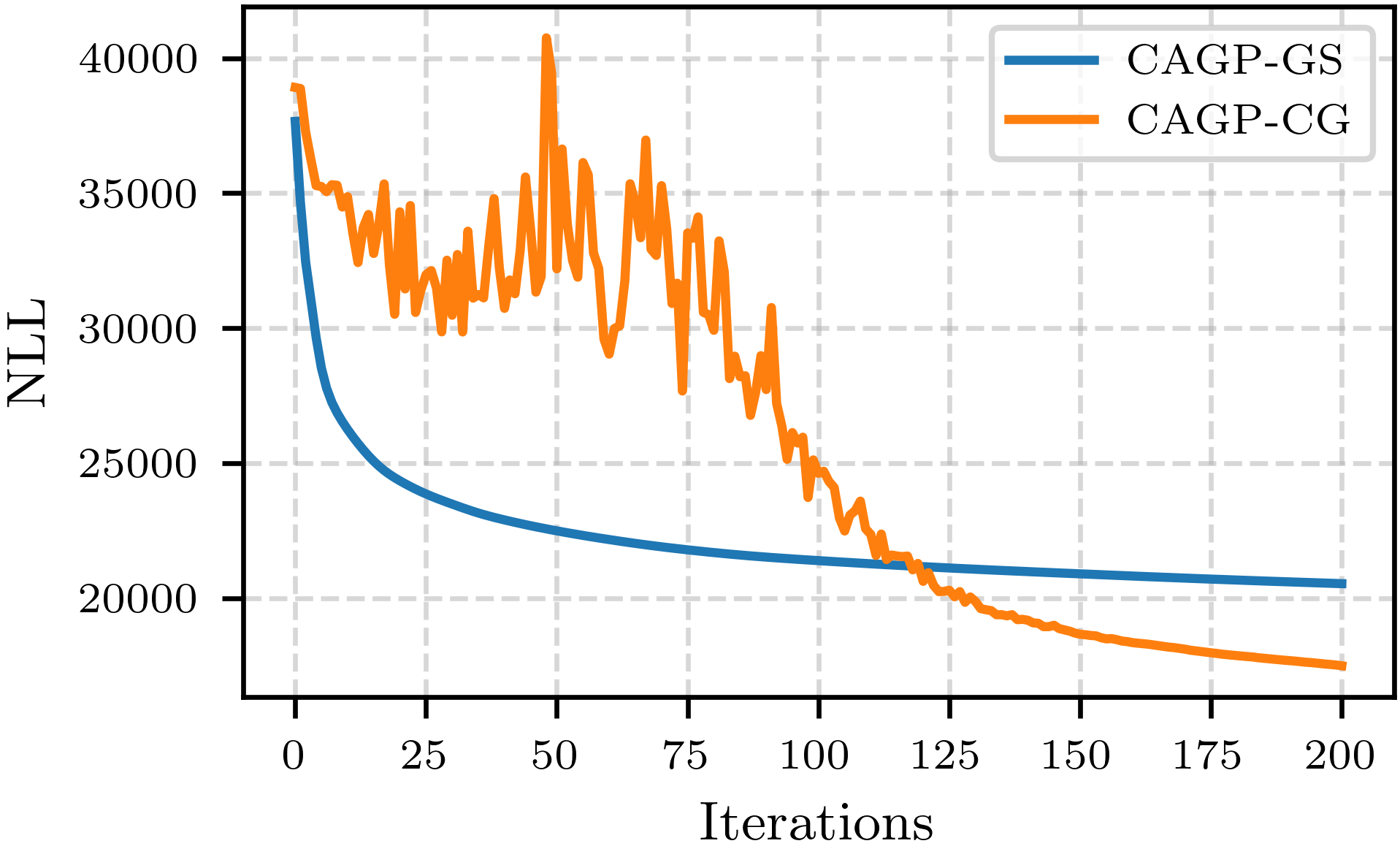}
        \caption{NLL for the large-scale regression from \cref{sec:era5}}
        \label{fig:era5:nll}
    \end{subfigure}
    \caption{RMSE and NLL for the large-scale regression from \cref{sec:era5}}
    \label{fig:era5:rmse_nll}
\end{figure*} 

\cref{fig:era5:rmse_nll} show the \ac{RMSE} and \ac{NLL} for this regression problem. As in \cref{sec:synthetic}, \ac{CAGP}-GS outperforms \ac{CAGP}-CG for inital iterations in both RMSE and NLL. However, unlike the case of synthetic experiment, \ac{CAGP}-GS outperforms \ac{CAGP}-CG in NLL for almost the same number of iterations as RMSE, but the difference in NLL is smaller. We speculate that this is due to model mismatch, since we are outside the synthetic ``in-model'' setting here unlike the previous section. 

\begin{figure*}
    \includegraphics[width=\textwidth]{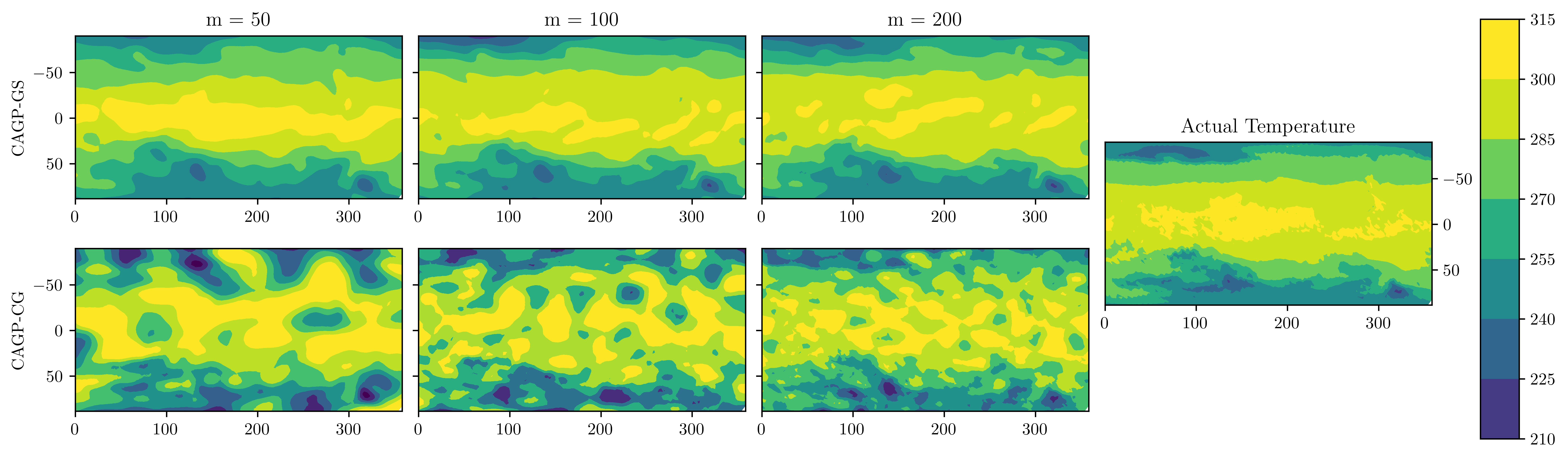}
    \caption{Sequences of posterior means for the large-scale regression from \cref{sec:era5}, as a function of $m$ for the detrended data. The x and y axes represent longitudes and latitudes respectively, and the contours indicate the temperature in Kelvin.}
    \label{fig:era5:mean_detrended} 
\end{figure*}

\begin{figure*}
    \begin{subfigure}{0.5\textwidth}
        \includegraphics[width=\textwidth]{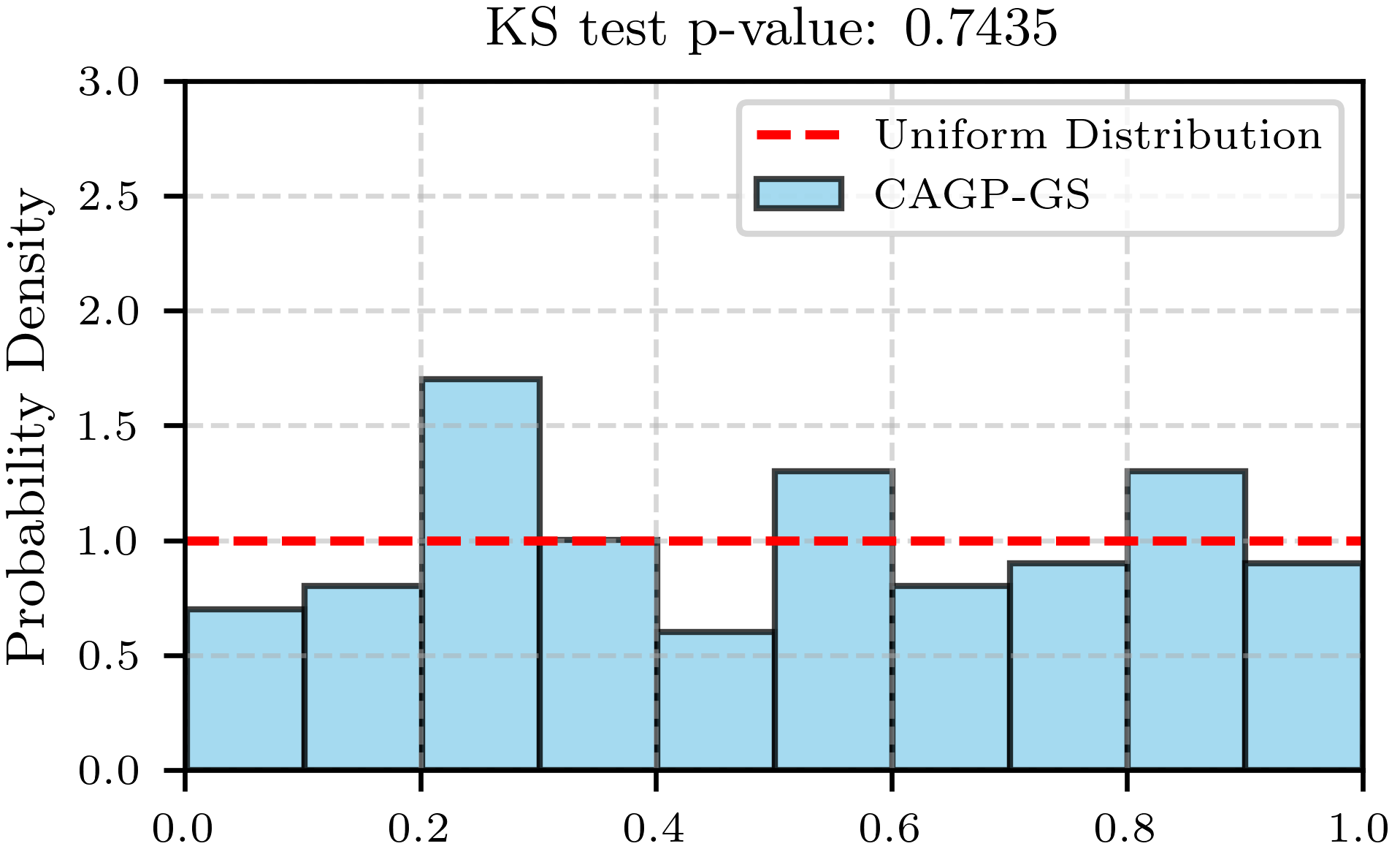}
        \caption{CAGP-GS}
    \end{subfigure}
    \begin{subfigure}{0.5\textwidth}
        \includegraphics[width=\textwidth]{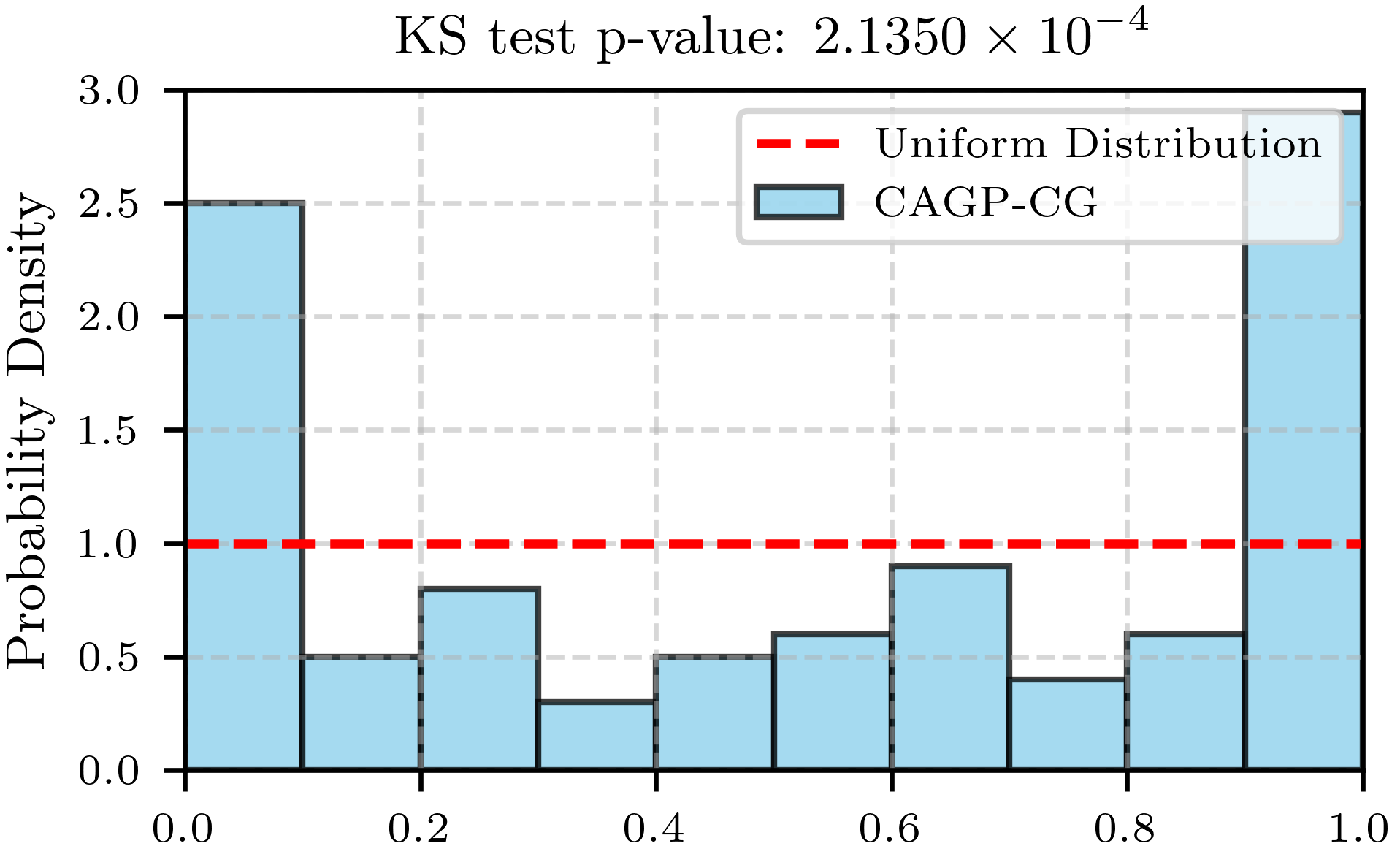}
        \caption{CAGP-CG}
    \end{subfigure}
    \caption{Calibration results for the large-scale regression problem from \cref{sec:era5} for the detrended data.} \label{fig:era5:zscores_detrended}
\end{figure*}

We test if the conclusions from \cref{sec:era5} is sensitive to trends in the data. As the global temperature is known to decrease as the latitude increases, using a detrended temperature data is a good way to examine this. We replace the constant prior mean in \cref{sec:era5} with a fitted quadratic polynomial of latitude. \cref{fig:era5:mean_detrended} shows the posterior mean for the settings same as \cref{fig:era5:mean}. As seen before, we see that the \ac{CAGP}-GS posterior mean is smoother than the \ac{CAGP}-CG posterior mean. \cref{fig:era5:zscores_detrended} shows the results of calibratedness experiment (for the same settings as \cref{fig:era5:zscores}). Again, the p-values from the Kolmogorov-Smirnov test for uniformity  show that \ac{CAGP}-GS is well-calibrated, while \ac{CAGP}-CG remains poorly calibrated. Detrending the data does not seem to affect the conclusions from before. 

\end{document}